\documentclass[runningheads]{llncs}

\usepackage{eccv}

\usepackage{eccvabbrv}

\usepackage{graphicx}
\usepackage{booktabs}
\usepackage{subcaption}
\usepackage[accsupp]{axessibility}  
\usepackage{multirow}
\usepackage[table]{xcolor}
\usepackage{amssymb}
\usepackage{pifont}

\usepackage{hyperref}

\usepackage{orcidlink}

\begin{document}

\title{TerraDiT: Point-Conditioned Diffusion Transformer for Satellite Image Synthesis} 

\titlerunning{TerraDiT}

\author{Srikumar Sastry\inst{*} \and
Dan Cher\inst{*}\and
Brian Wei\inst{*} \and Aayush Dhakal \and Subash Khanal \and Dev Gupta \and Nathan Jacobs}

\authorrunning{Sastry et al.}

\institute{Washington University, St. Louis MO 63130, USA\\
\email{\{s.sastry, cher, b.j.wei, a.dhakal, k.subash, dev.g, jacobsn\}@wustl.edu}\\
*Equal Contribution}

\maketitle

\begin{abstract}
We introduce TerraDiT, a diffusion transformer designed for text-to-satellite image generation with point-based control. Existing controlled satellite image generative models often require pixel-level maps that are time-consuming to acquire, yet semantically limited. To address this limitation, we introduce a novel point-based conditioning framework that controls the generation process through the spatial location of the points and the textual description associated with each point, providing semantically rich control signals. This approach enables flexible, annotation-friendly, and computationally simple inference for satellite image generation. To this end, we introduce an adaptive local attention mechanism that effectively regularizes the attention scores based on the input point queries. We systematically evaluate various domain-specific design choices for training TerraDiT, including the selection of satellite image representation for alignment and geolocation representation for conditioning. Our experiments demonstrate that TerraDiT achieves impressive generation performance, surpassing the state-of-the-art remote sensing generative models. Our models, dataset, and code are available at \url{https://github.com/mvrl/TerraDiT}.
\end{abstract}

\section{Introduction}
\label{sec:intro}

The recent success in advancing text-to-image generation can be attributed to improvements in architectures, efficient training algorithms, and large-scale computational infrastructures. Diffusion Transformers~\cite{peebles2023scalable} (DiTs) and Flow Matching~\cite{lipman2022flow} (FM) represent two significant advancements in this field. DiTs, in conjunction with FM, have been increasingly utilized in various generative modeling pipelines, such as SD3~\cite{esser2024scaling}, PixArt-$\alpha$~\cite{chen2023pixart}, Sora-2~\cite{sora} and Wan-2.1~\cite{wan2025wan}, which have outperformed traditional UNet-based diffusion models. Frameworks such as LightningDiT~\cite{yao2025reconstruction} and REPA~\cite{yu2024representation} have provided further improvements in training efficiency.

\begin{figure}[!h]
  \centering
  \includegraphics[width=\linewidth]{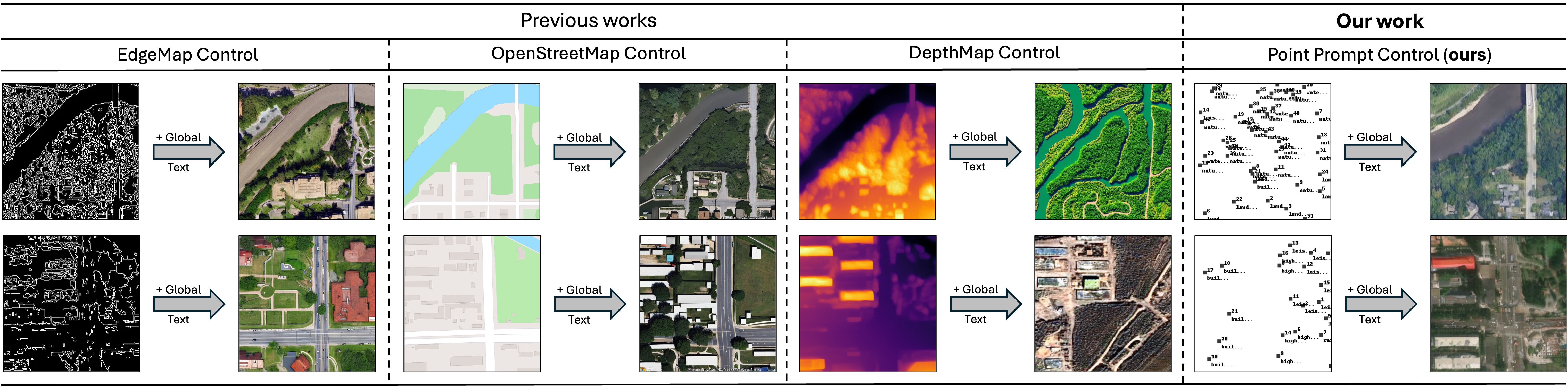}
  \caption{Existing methods for satellite image synthesis rely on dense spatial layout controls that are expensive and time-consuming to acquire, yet semantically limited. Our proposed method enables conditioning the generation process using semantically rich point queries, where each point is associated with a free-form text prompt. The spatial layout of the generation is guided by the point locations, while the semantics are driven by the accompanying text. Point queries do not impose strict pose and shape constraints on the generation process, resulting in a wide variety of semantically consistent satellite images (see Figure~\ref{point_gen}).}
  \label{img:teaser}
\end{figure}

However, this success has not been well explored in specialized generative modeling pipelines, such as those used in remote sensing, which require careful domain-specific design choices. In contrast, visual representation learning for satellite imagery has experienced rapid advancements through frameworks like AnySat~\cite{astruc2025anysat} and DINOv3~\cite{simeoni2025dinov3}. Nonetheless, the representations generated by these powerful models have not been investigated for satellite image generation. To address this gap, we propose satellite representation alignment, which builds upon REPA~\cite{yu2024representation} by aligning the hidden representations of DiTs with strong satellite image encoders.

While text-to-satellite image generative models provide flexibility, their applications are somewhat limited. In contrast, satellite image generation with spatial layout control has emerged as a crucial component in diverse applications, including urban design~\cite{sastry2024geosynth}, environmental monitoring~\cite{goktepeecomapper}, and data augmentation~\cite{toker2024satsynth}. As illustrated in Figure~\ref{img:teaser}, existing controlled satellite image generative models require dense, pixel-level maps, which can be costly to acquire during inference. Furthermore, these pixel-level maps are often semantically limited, representing a fixed set of concepts. To this end, we propose a novel framework that enables controlling the generation process with semantically rich point queries, where each point is associated with a free-form textual description. To achieve this, we design an adaptive local attention module that effectively allows kernel-weighted localized cross-attention for point-based conditions.

\begin{figure}[!t]
  \centering
  \includegraphics[width=0.6\linewidth]{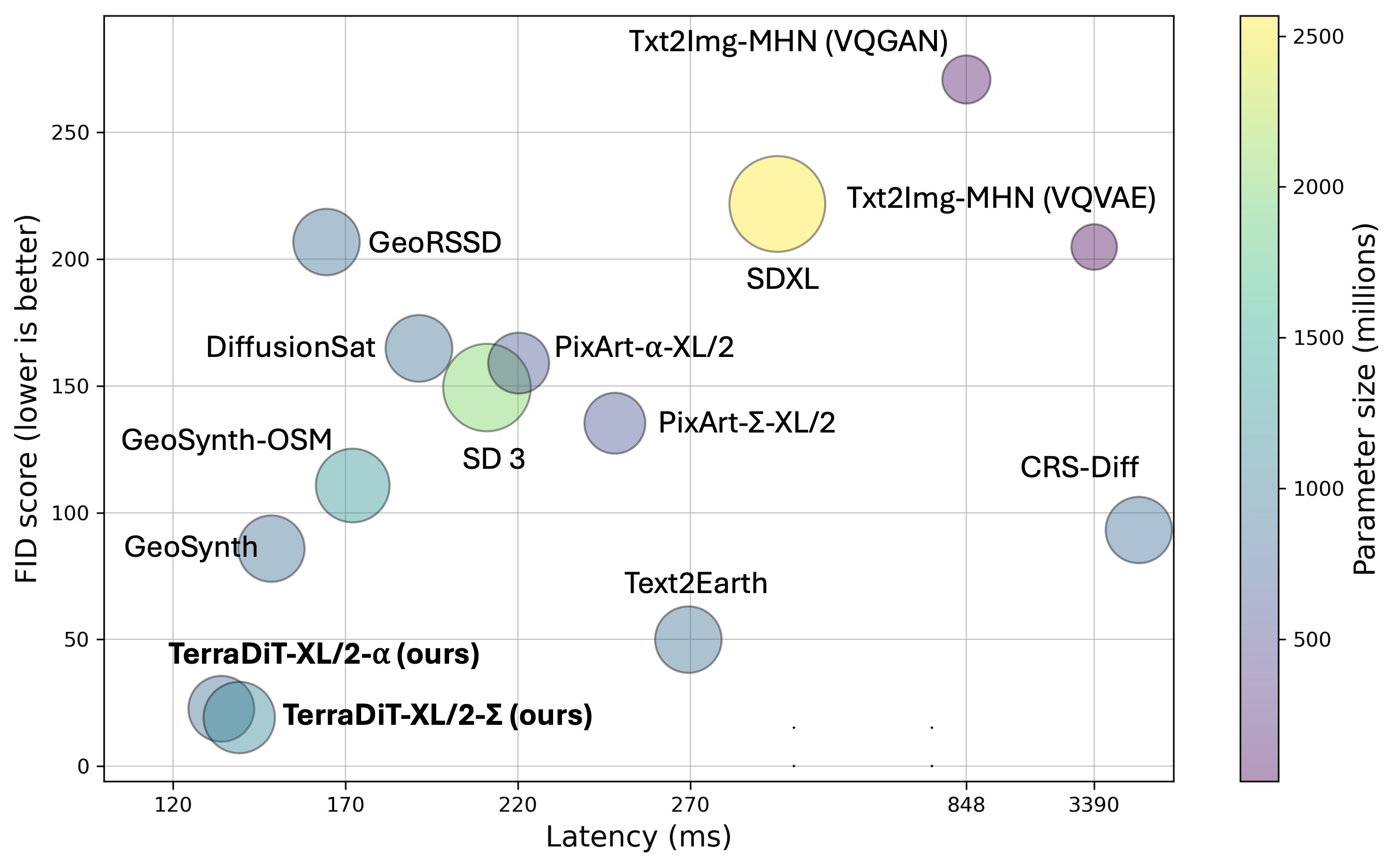}
  \caption{\textbf{Performance vs Latency.} We compare the performance of several remote sensing generative models against the time taken to generate a single image. Both variants of our proposed model, TerraDiT, are efficient and outperform the state-of-the-art generative models.}
  \label{fig:bubble}
\end{figure}

We introduce TerraDiT, a satellite image generative model that significantly outperforms UNet-based models while reducing computational demands. TerraDiT is trained in two variants: TerraDiT-XL/2-$\alpha$, a text-to-satellite image generative model, and TerraDiT-XL/2-$\Sigma$, a point-controlled satellite image generative model. As shown in Figure~\ref{fig:bubble}, both our models outperform state-of-the-art satellite image generation models and achieve the lowest latency in single image generation. The contributions of our work are as follows:
\begin{enumerate}
    
    \item We introduce a DiT model, TerraDiT-XL/2-$\alpha$, that achieves state-of-the-art text-to-satellite image generation and provides a strong foundation for controlled remote sensing synthesis.
    \item We introduce a point-conditioned DiT, TerraDiT-XL/2-$\Sigma$, that enables fine-grained spatial and semantic controlled generation using minimal point-based inputs, offering a flexible alternative to dense pixel-level supervision.
    \item We propose an adaptive local attention (ALA) module that regularizes cross-attention between latent image tokens and point prompts, resulting in more precise and spatially coherent conditioning.
    \item We systematically study key design factors for satellite image diffusion transformers, including conditioning strategies, geolocation encoders, and representation alignment to satellite image encoders.
\end{enumerate}

\section{Related Work}
\subsection{Controlled Image Generation}
Base diffusion models typically rely on global text prompts for guidance, which limits their ability to provide fine-grained spatial control. To address this, earlier works~\cite{yang2023reco, avrahami2023spatext, bashkirova2023masksketch} introduced explicit spatial annotations as conditioning signals in the form of bounding boxes, segmentation maps, and free-form sketches. Although effective, these approaches are restricted to specific annotation types and struggle to generalize across control modalities.

Universal control architectures were developed to handle multiple conditioning signals within a single framework. ControlNet~\cite{zhang2023adding} introduced trainable control branches compatible with pre-trained diffusion backbones, inspiring follow-ups~\cite{zhao2023uni, li2023gligen, chen2024pixartdelta}, which reduce training overhead while supporting diverse inputs. Other works~\cite{huang2023composer, zheng2023layoutdiffusion, lee2024groundit} began to unify compositional and layout control, paving the way for multi-condition generation frameworks.

Recent works have continued this trajectory toward more unified, fine-grained, and efficient control. Some models~\cite{peng2024controlnext, li2024controlnet++} improve parameter efficiency and condition-output consistency, while others~\cite{wang2024instancediffusion} introduce instance-level control over object placement and attributes. Unified frameworks~\cite{gandikota2024unified, mu2025editar, wang2024omnicontrolnet, tan2025ominicontrol, xiao2025omnigen} have further collapsed multi-task control into a single architecture. While these works have simplified architectural inputs, generating fine-grained and localized edits within complex scenes has recently come into focus~\cite{zhao2025local, choi2025finecontrolnet, wu2025ifadapter}. While these advances have enabled remarkable control over image generation, they rely heavily on dense spatial controls. Building on this progression toward precise and unified control, we take a step toward fine-grained generation in sparse settings where dense control signals are impractical.

\subsection{Diffusion Transformers}
The diffusion transformer (DiT) replaces the convolutional U-Net with a transformer backbone in a latent diffusion model, using zero-initialized adaptive layer normalization for stable conditioning and scalable training~\cite{peebles2023scalable}. PixArt-$\alpha$~\cite{chen2023pixart}, and PixArt-$\Sigma$~\cite{chen2024pixart} incorporate text-based cross-attention, and Stable Diffusion 3~\cite{esser2024scaling} enables bidirectional text-image interaction to improve text-guided and multimodal image generation. Complementary efforts such as SiT~\cite{ma2024sit} reformulate the generative process itself, unifying flow- and diffusion-based modeling within a single DiT backbone and systematically evaluating key architectural design choices. Several DiT variants refine the internal denoising process to better capture both global structure and local detail. Dynamic-DiT~\cite{jia2025d} introduces a Dynamic VAE with multi-granularity noise prediction to unify coarse and fine spatial information, while DiffiT~\cite{hatamizadeh2024diffit} employs time-dependent attention to adapt the denoising behavior across timesteps. Several variants aim to make DiTs faster and more stable to train: MaskDiT~\cite{zheng2023fast} leverages masked-image modeling, SD-DiT~\cite{zhu2024sd} incorporates a joint generative–discriminative objective, FasterDiT~\cite{yao2024fasterdit} improves SNR weighting and introduces velocity-direction supervision, and LightningDiT~\cite{yao2025reconstruction} aligns the VAE with a vision foundation encoder to enhance reconstruction and stability. REPA~\cite{yu2024representation} took a complementary approach by regularizing training through the alignment of noisy diffusion representations with pretrained visual encoders. Building on these developments, TerraDiT extends the transformer-based diffusion paradigm to the remote-sensing domain through representation alignment and novel point-based conditioning for semantically grounded and spatially controllable satellite image generation.

\subsection{Geospatial Generative Models}

The field of satellite image generation has rapidly evolved with the development of several text-conditioned satellite image generative models~\cite{xu2023txt2img,zhang2024rs5m,yu2024metaearth,liu2025text2earth}, with recent work exploring various conditions tailored to remote sensing imagery. Unlike natural image generation where text-only conditioning is common, satellite image generation has adopted various paradigms to improve semantic coherence and spatial control. DiffusionSat~\cite{khanna2023diffusionsat} modeled imagery evolution over time with temporal and metadata conditioning. Cross-View Meets Diffusion~\cite{arrabi2025cross} synthesized ground-to-aerial (G2A) imagery using GPG2A, conditioning birds-eye view generation on street-level imagery.

Beyond these multi-modal approaches, the dominant paradigm has been dense spatial control through pixel-level semantic layouts. CRS-Diff~\cite{tang2024crs}, GeoSynth~\cite{sastry2024geosynth}, and EarthSynth~\cite{pan2025earthsynth} enable the use of segmentation maps or OpenStreetMap (OSM) layouts to guide generation. While providing fine-grained control, these approaches require dense annotations that are costly to acquire, yet are semantically limited. Recent efforts have explored more efficient conditioning, with CC-Diff++~\cite{zhang2025cc} introducing automatic layout prediction from text descriptions, where bounding boxes guide spatial arrangement. 

Despite these advances, existing models have primarily relied on UNet-based architectures. While diffusion transformers~\cite{peebles2023scalable} and Flow Matching~\cite{lipman2022flow} have proven effective for natural images, their application to satellite imagery remains underexplored. We combine Diffusion Transformers with domain-specific representation alignment and propose a novel point-based conditioning framework that enables flexible spatial and semantic guidance through sparse point queries with textual descriptions, maintaining fine-grained control while significantly reducing annotation requirements.

\section{Method}
\subsection{Preliminaries}
Flow-based generative models~\cite{lipman2022flow, liu2022flow, albergo2023stochastic} learn to gradually turn noise $\epsilon \sim \mathcal{N}(0, \textbf{I})$ into the data distribution $\hat{\text{\textbf{x}}}\sim p(\text{\textbf{x}})$, by predicting a velocity field \textbf{v}($\text{\textbf{x}}_t, t$) that points from $\epsilon$ to $\hat{\text{\textbf{x}}}$. This process can be represented using a continuous time equation:
\begin{equation}
    \text{\textbf{x}}_t = \alpha_t\hat{\text{\textbf{x}}} + \sigma_t\epsilon
\end{equation}
where $\alpha_t$ and $\sigma_t$ are time-dependent drift and diffusion coefficients, respectively. This can be modeled as a probability flow ODE with the velocity field as follows:
\begin{equation}
    \dot{\text{\textbf{x}}}_t =  \textbf{v}(\text{\textbf{x}}_t, t).
\end{equation}
Solving this equation starting from Gaussian noise leads to samples in the data distribution $p(\text{\textbf{x}})$. The velocity can be estimated using a neural network $\textbf{v}_{\theta}(\text{\textbf{x}}_t, t)$ and minimizing the following objective:
\begin{equation}
    \mathcal{L}_\textbf{v} = \mathop{\mathbb{E}}_{\hat{\text{\textbf{x}}}, \epsilon,t}[||\textbf{v}_{\theta}(\text{\textbf{x}}_t, t) - \dot{\alpha_t}\hat{\text{\textbf{x}}} - \dot{\sigma_t}\epsilon||^2].
\end{equation}

Diffusion transformers~\cite{peebles2023scalable} are a class of transformer-based architectures designed for diffusion and flow-based modeling. These architectures operate on image patches as tokens, providing the flexibility to operate in both the pixel and latent spaces. DiTs often employ Adaptive LayerNorm (AdaLN) to condition the generation process with vector embeddings. AdaLN has demonstrated superior performance to standard vanilla cross-attention for conditioning on diffusion timesteps and additional vector-based embeddings. The final layer in a DiT decodes the tokens back into the input space, essentially reversing the patching operation. When performing diffusion or flow in the latent space, the DiT output is passed to a pixel decoder (e.g., a VAE decoder), reconstructing the final image.

\subsection{Dataset Construction}
Our dataset builds upon the Git-10M dataset~\cite{liu2025text2earth}, which offers high-resolution satellite imagery paired with image-level text descriptions generated by GPT-4o~\cite{achiam2023gpt4}. For this work, we focus on high-resolution satellite imagery at a zoom level of 17, with a ground sampling distance of approximately 1 meter. The resulting dataset contains 2 million images. To enable grounded point-level conditioning, we augment each image with polygon-level OpenStreetMap (OSM) vector data obtained from GeoFabrik extracts~\cite{geofabrik_osm}. Following ~\cite{cher2026vectorsynth}, we generate co-registered OSM vector tiles that are spatially aligned with the satellite images. During training, we randomly sample 10 to 50 points from a vector tile, extracting both the corresponding OSM tag text and relative pixel coordinates. This localized grounded conditioning is provided to the model.

To evaluate our generative models, we created two separate testing splits of the Git-10M dataset. Git-Rand-15k is constructed by randomly selecting 15k samples from the dataset. On the other hand, Git-Spatial-15k is carefully selected to ensure that the samples are geographically held out from the rest of the training dataset. This allows us to evaluate the geospatial adaptation of our models. These two evaluation splits enable us to effectively assess our models’ ability to capture two distinct characteristics of satellite image distributions. The spatial distribution of our training and evaluation datasets is presented in the appendix. 

\subsection{TerraDiT}
\begin{figure}[!t]
    \centering
    \begin{subfigure}[t]{0.42\textwidth}
        \centering
        \includegraphics[width=\linewidth]{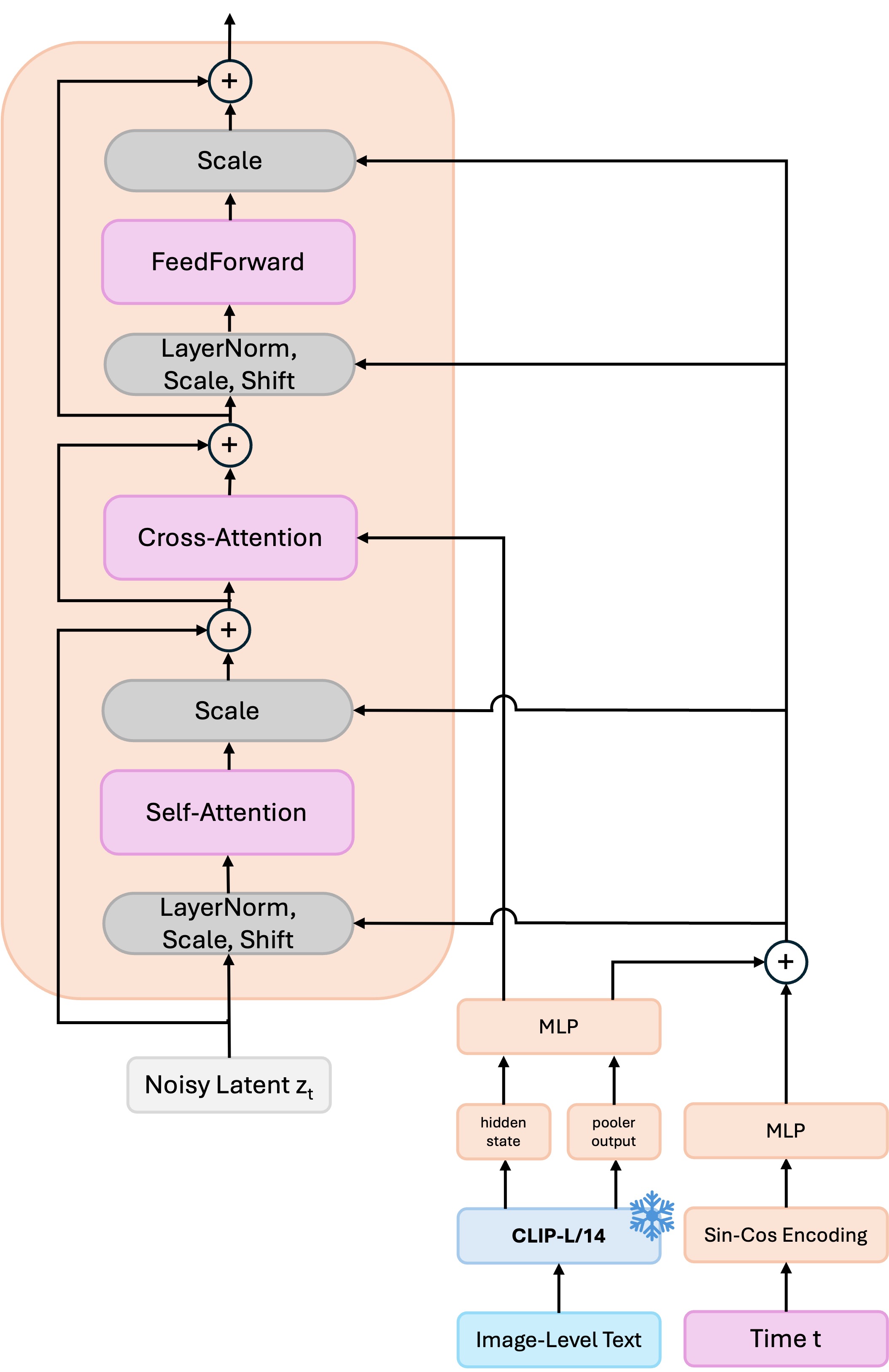}
        \caption{Single block of our TerraDiT-XL/2-$\alpha$ model. It shows a single block of our text-to-satellite image generative model.}
        \label{fig:alpha}
    \end{subfigure}
    \hfill
    \begin{subfigure}[t]{0.53\textwidth}
        \centering
        \includegraphics[width=\linewidth]{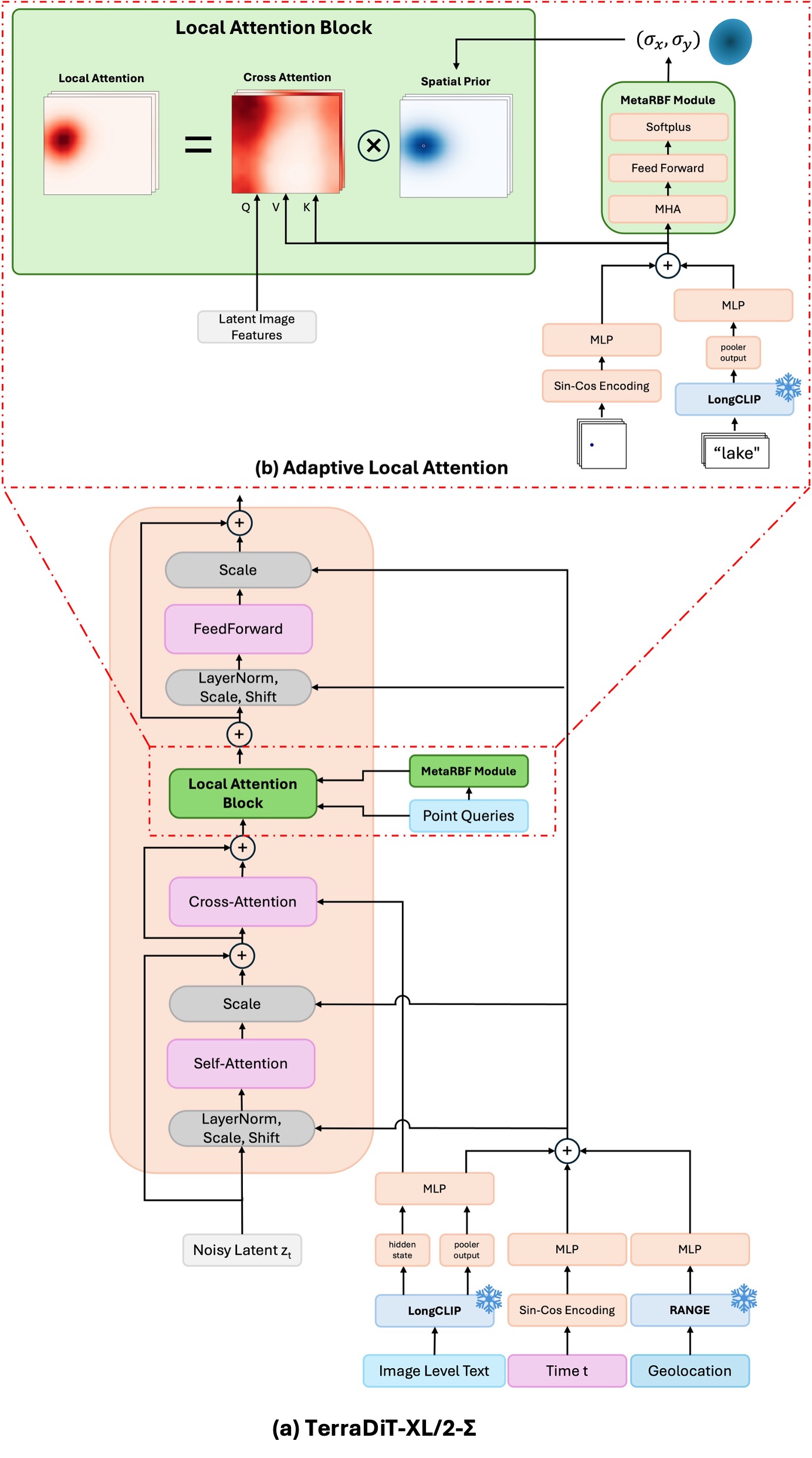}
        \caption{Single block of our TerraDiT-XL/2-$\Sigma$ model. It shows (a) a single block of our point-controlled satellite image generative model, and (b) our proposed Adaptive Local Attention (ALA) block.}
        \label{fig:sigma}
    \end{subfigure}
    \caption{Proposed architecture of our TerraDiT-XL/2-$\alpha$ (left) and TerraDiT-XL/2-$\Sigma$ (right) models.}
\end{figure}

We follow the DiT architecture as employed in SiT~\cite{ma2024sit}. To minimize computational requirements, we adopt a three-stage training procedure. Initially, we train an unconditional model that learns the overall distribution of satellite images within our training dataset. Subsequently, we introduce text conditioning and train for the task of text-to-satellite image generation. As depicted in Figure~\ref{fig:alpha}, we incorporate a single MultiHead Cross-Attention layer within each of our DiT blocks. The output of this cross-attention layer is initialized to zero to prevent noisy gradient updates. We utilize LongCLIP~\cite{zhang2024long} as our text encoder and condition our model with both the dense tokens and single embedding representation generated by the text encoder. The single embedding representation is added to the flow timestep embedding. The dense tokens from the encoder are then passed to the cross-attention module within the DiT block.

In the third stage, we train for the task of point-controlled satellite image generation. Each point query is associated with a 2D coordinate within the image and a textual prompt. The coordinates are encoded using 2D sine-cosine encoding followed by an MLP. Similarly to stage two, we use LongCLIP\footnote{\tiny https://huggingface.co/zer0int/LongCLIP-KO-LITE-TypoAttack-Attn-ViT-L-14}~\cite{zhang2024long} to encode each textual prompt. The embeddings are added element-wise and passed to our model. In this stage, we introduce our Adaptive Local Attention (ALA) block, as illustrated in Figure~\ref{fig:sigma}. We provide additional details about our ALA block in the following section. We further add geolocation conditioning through adaptive layer normalization, leveraging RANGE~\cite{dhakal2025range} for geolocation-aware embeddings. Inspired by REPA~\cite{yu2024representation}, during each training stage, we align the hidden representations from our model with the satellite-only DINOv3~\cite{simeoni2025dinov3} model. Specifically, we learn a linear projection from the hidden representation of a single TerraDiT block to the DINOv3 representation space. For geolocation conditioning, we find RANGE~\cite{dhakal2025range} embeddings to perform best. Ablations evaluating all design choices along with exact details about our architecture are provided in the appendix.

\begin{figure}[!ht]
    \centering
    \includegraphics[width=0.5\linewidth]{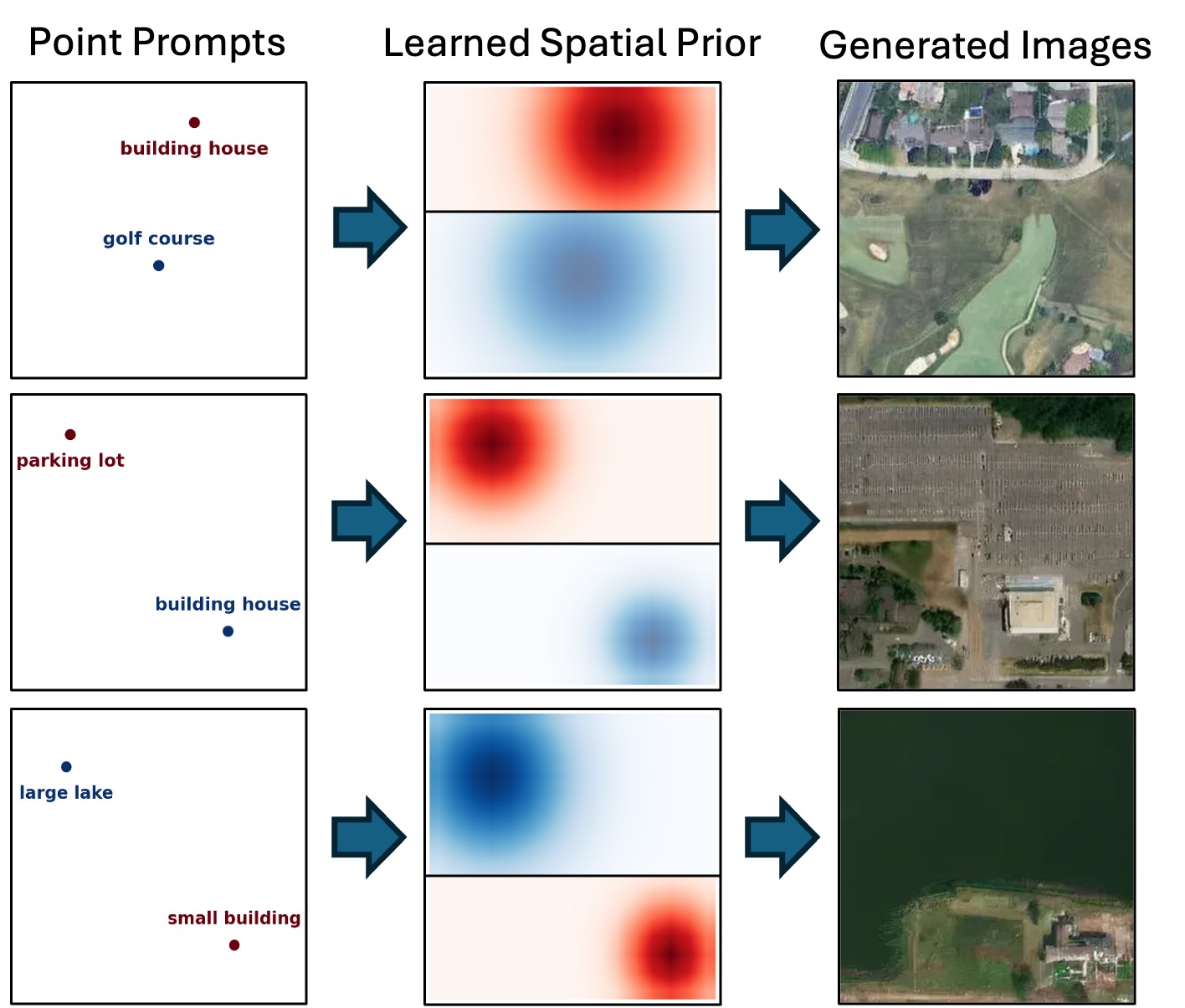} 

    \caption{\textbf{Learned spatial prior for various concepts}. Notice that the model has learned concept specific spatial priors such as for ``small building" predicting a smaller spatial extent.}
    \label{fig:asla_viz}
\end{figure}

\subsection{Adaptive Local Attention}
We propose ALA, a localized attention mechanism that conditions our model with point-based controls. As illustrated in Figure~\ref{fig:sigma} (b), ALA comprises two modules: 1) the MetaRBF module, which predicts the spatial extent of each point-specified concept and 2) the Local Attention Block, which uses these predictions to modulate attention. The MetaRBF module serves as a spatial prior that determines how strongly each concept should influence each image token. It comprises a MultiHead Attention (MHA) block followed by a feedforward layer and a softplus activation. The MetaRBF module predicts $\sigma_\text{x}$ and $\sigma_\text{y}$ values corresponding to each point query, which determine the spatial extent of each concept in the x and y directions, respectively. These $\sigma$ values parameterize a 2-dimensional Radial Basis Function (RBF) centered at the location specified by each point query. Let $\mathcal{P} = \{(x_1, y_1), (x_2, y_2) \ldots, (x_n, y_n)\}$ denote the spatial location of points within the image canvas. Similarly, let $\mathcal{S} = \{(\sigma_\text{x}^1, \sigma_\text{y}^1), (\sigma_\text{x}^2, \sigma_\text{y}^2) \ldots, (\sigma_\text{x}^n, \sigma_\text{y}^n)\}$ denote the set of extents predicted by our MetaRBF module for each point. The RBF kernel is then modeled as follows:
\begin{equation}
    p_s(i, j) = \text{exp} \left\{- \frac{(j_x - x_i)^2}{(\sigma_\text{x}^i)^2} - \frac{(j_y - y_i)^2}{(\sigma_\text{y}^i)^2}\right\}
 \end{equation}
where $j_x$ and $j_y$ are 2D position of the $j^{\text{th}}$ latent token and  $p_s(i, j)$ is the likelihood value of our spatial prior corresponding to the $i^{\text{th}}$ point query and $j^{\text{th}}$ latent token. These spatial prior matrices modulate the cross attention operation between the latent queries and the point prompts keys and values in the Local Attention Block. This modulation allows the model to attend locally around each point query while respecting the spatial extents of each concept predicted by the MetaRBF module.

\begin{table*}[!ht]
\centering
\resizebox{0.9\linewidth}{!}{
\begin{tabular}{l|llllll}
Variant & Params (M) & Epochs & Training Steps & GPU Days & Latency (ms) & GFLOPS \\
\toprule
Unconditional & 682.61 & 102 & 800k & 11.88 & 124.52 & 2182.07 \\
Text & 836.69\textsuperscript{\textcolor{gray}{+154.08}} & 121\textsuperscript{\textcolor{gray}{+19}} & 950k\textsuperscript{\textcolor{gray}{+150k}} & 19.69\textsuperscript{\textcolor{gray}{+7.81}} & 134.00\textsuperscript{\textcolor{gray}{+9.48}} & 3661.51\textsuperscript{\textcolor{gray}{+1480.44}} \\
Text+Points+Geolocation & 1109.89\textsuperscript{\textcolor{gray}{+273.20}} & 127\textsuperscript{\textcolor{gray}{+6}} & 1M\textsuperscript{\textcolor{gray}{+50k}} & 22.49\textsuperscript{\textcolor{gray}{+2.80}} & 139.23\textsuperscript{\textcolor{gray}{+5.23}} & 7316.47\textsuperscript{\textcolor{gray}{+3654.96}} \\
\end{tabular}
}
\caption{Information about each training stage for our TerraDiT models. Numbers in superscript (grey) indicate the incremental addition introduced at each stage. Latency is measured for single image generation with 28 steps at half precision.}
\label{tab:training_days}
\end{table*}

\begin{figure*}[!b]
  \centering
  \includegraphics[width=\linewidth]{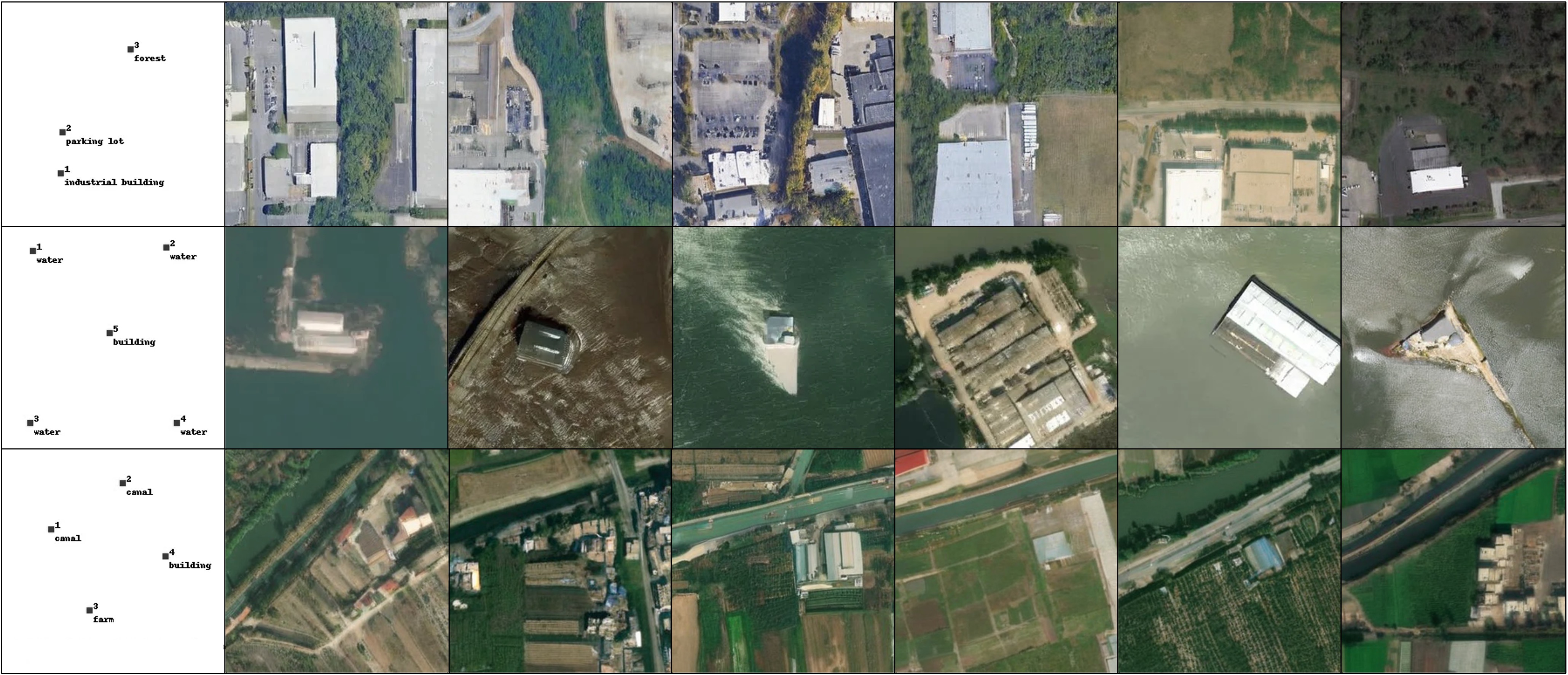}
  \caption{\textbf{Samples from TerraDiT-XL/2-$\mathbf{\Sigma}$}. Controlling the generation process with point queries without a global text prompt enables flexible and diverse satellite image generation without strict pose and shape constraints. In particular, as shown in the last row, our model generates a single consistent canal from just two input points.}
  \label{point_gen}
\end{figure*}

\begin{figure}[!h]
    \centering
    \includegraphics[width=0.6\linewidth]{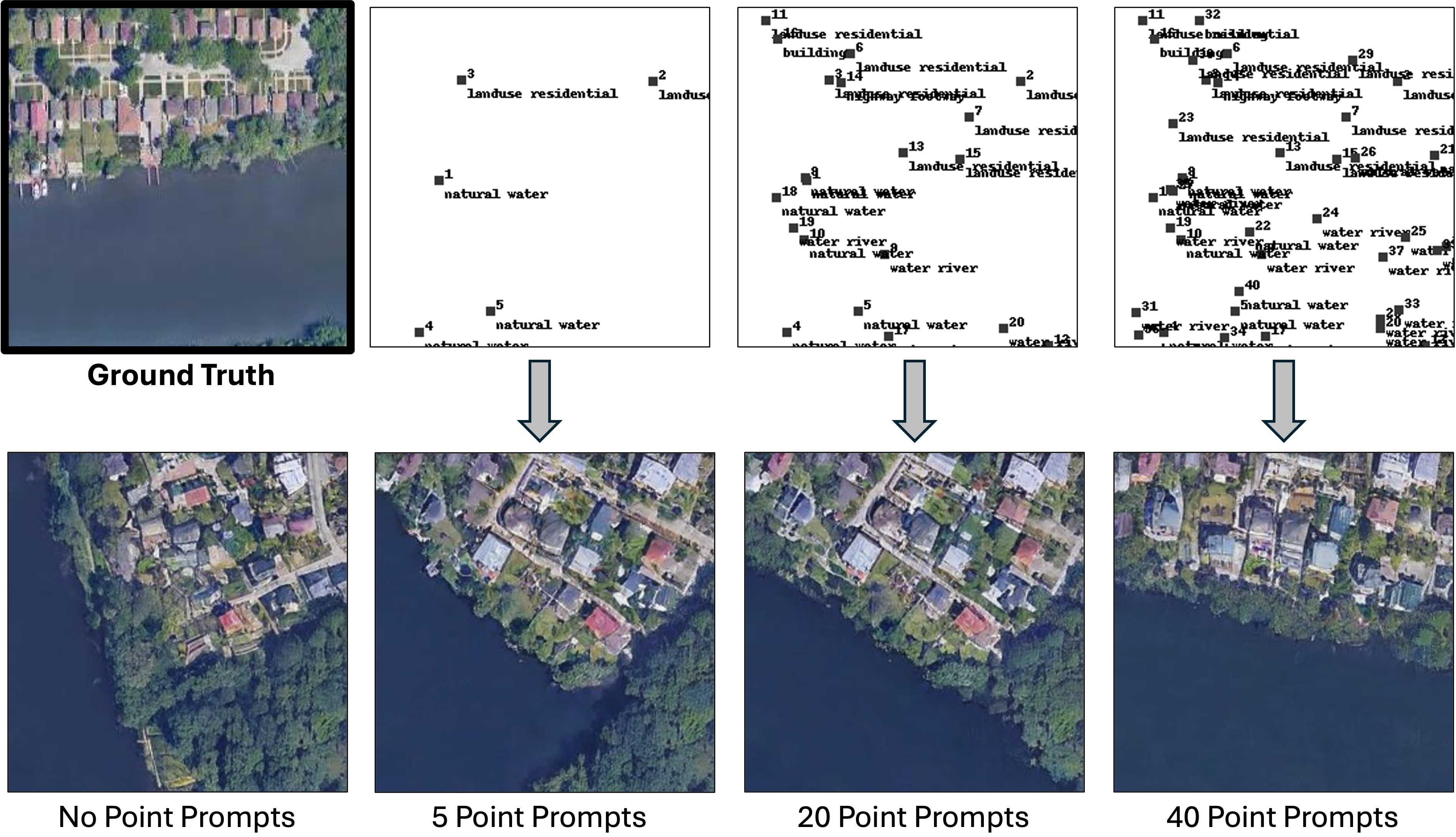} 

    \caption{\textbf{Varying point queries}. We generate images with a varying number of input point queries. It shows that our model increases the consistency of the generated scene with the ground truth on increasing points.}
    \label{fig:pointwise}
\end{figure}

\begin{figure*}[!ht]
  \centering
  \includegraphics[width=\linewidth]{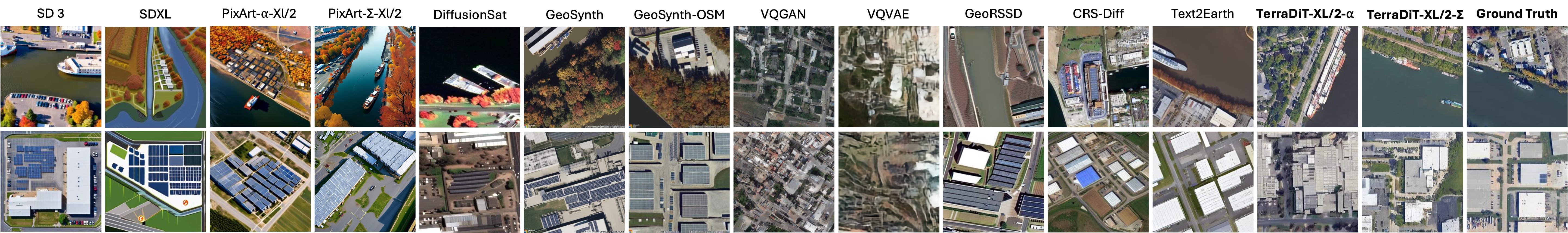}
  \caption{\textbf{Visual Quality of Generation}. Satellite imagery generated by various state-of-the-art generative models compared with TerraDiT and ground truth. Natural image generative models have low perceptual quality while satellite image generative models have poor structural consistency with the ground truth.}
  \label{all_comp}
\end{figure*}

\begin{figure*}
  \centering
  \includegraphics[width=\linewidth]{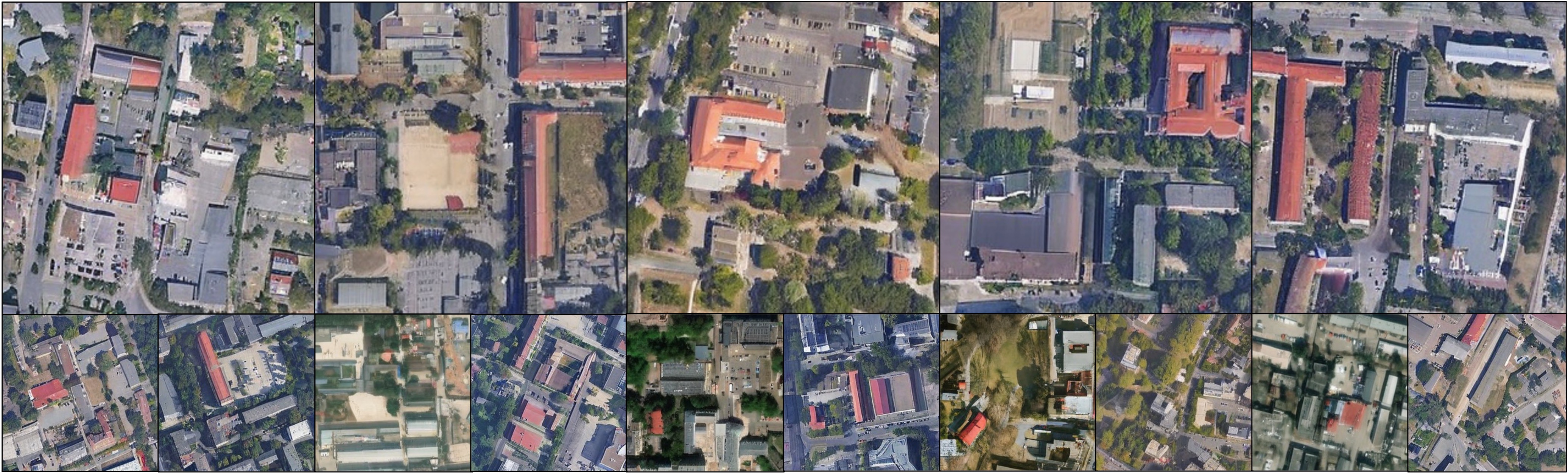}
  \caption{\textbf{Samples from TerraDiT-XL/2-$\mathbf{\alpha}$}. The images were generated using the caption: ``This satellite image depicts an urban area with several buildings and structures. There are multiple rectangular buildings, some with red or gray roofs. A parking lot with numerous vehicles is visible in the center of the image. Trees and vegetation are scattered throughout the area, particularly around the buildings.''}
  \label{collage:geodit-alpha}
\end{figure*}

\section{Experiments}
For all comparisons, we train TerraDiT-XL/2, the largest variant of DiT under consideration. It consists of 28 DiT blocks with a hidden dimension of 1152 and a patch size of 2. We train our unconditional model for a total of 800k steps. We then train our text-to-satellite generation model for an additional 150k steps. Finally, we train our point-conditioned model for an additional 50k steps. In total, we trained our model for 1M steps. See Table~\ref{tab:training_days} for additional information. All the experiments are conducted on 4 NVIDIA-H100 GPUs with an effective batch size of 256, AdamW optimizer and learning rate of 1e-5. In this section, we provide several experiments and ablation studies to evaluate our models’ performance. Additional experiments and qualitative results are provided in the appendix.

\subsection{Qualitative Evaluation}
We present qualitative visualizations of satellite imagery generated using our models. Figure~\ref{point_gen} showcases results from our TerraDiT-XL/2-$\Sigma$ model. We provide arbitrary point queries, each associated with a textual concept, without a global textual prompt or geolocation. As shown in the figure, our model generates high-fidelity satellite imagery that accurately represents the point-query semantics. Since our model's point queries do not impose strict pose and layout constraints, our model can generate a wide variety of satellite imagery that is semantically consistent for the same input. This contrasts with models using map-based controls, where the map input limits the shape of various objects within a scene. We see that the spatial layout of the generated scene improves with an increasing number of queries, confirming that point queries are an effective spatial control mechanism (see Figure~\ref{fig:pointwise}).

We compare the visual quality of satellite imagery generated by several state-of-the-art models, as shown in Figure~\ref{all_comp}. Models not specifically trained on satellite imagery, such as SD3, SDXL, and PixArt, often produce images with lower perceptual fidelity, although they generally respect the textual conditions. Text2Earth and GeoSynth generate images that are both perceptually realistic and semantically meaningful, while GeoSynth-OSM further improves spatial grounding, producing images more closely aligned with the ground truth. Our models consistently outperform these baselines, achieving high perceptual quality while faithfully representing the input semantics. The point-controlled variant, in particular, produces images that are visually realistic and better structurally aligned with the ground truth.

In Figure~\ref{collage:geodit-alpha}, we showcase several examples of satellite imagery generated by our TerraDiT-XL/2-$\alpha$ model. The images are remarkably well-aligned with the input caption. Notably, the images accurately depict “red and gray roof buildings,” as specified in the caption. Perceptually, the images appear realistic and of high quality. Additional visualizations are provided in the appendix.

\begin{table*}[!ht]
\centering
\resizebox{\linewidth}{!}{
\begin{tabular}{l|c|cccc|cccc|cccc|cccc}
\multirow{2}{*}{Model} & \multirow{2}{*}{Condition} & 
\multicolumn{4}{c|}{\textbf{Git-Rand-15k}} & 
\multicolumn{4}{c|}{\textbf{Git-Spatial-15k}} &
\multicolumn{4}{c|}{\textbf{FMoW}~\cite{christie2018functional}} &
\multicolumn{4}{c}{\textbf{RSICD}~\cite{lu2017exploring}} \\
\cline{3-18}
 &  & FID $\downarrow$ & LPIPS $\downarrow$ & SSIM $\uparrow$ & CLIP $\uparrow$ &
FID $\downarrow$ & LPIPS $\downarrow$ & SSIM $\uparrow$ & CLIP $\uparrow$ &
FID $\downarrow$ & LPIPS $\downarrow$ & SSIM $\uparrow$ & CLIP $\uparrow$ &
FID $\downarrow$ & LPIPS $\downarrow$ & SSIM $\uparrow$ & CLIP $\uparrow$ \\
\toprule
SDXL~\cite{podell2023sdxl} & $T$ & 150.10 & 0.5929 & 0.1817 & 0.2423 & 221.76 & 0.5845 & 0.2597 & 0.2389 & 131.60 & 0.6315 & 0.1288 & 0.2396 & 111.93 & 0.5910 & 0.1763 & 0.2469 \\
SD 3~\cite{esser2024scaling} & $T$ & 90.45 & 0.5083 & 0.1598 & 0.2567 & 149.34 & 0.5353 & 0.2000 & 0.2691 & 90.00 & 0.6063 & 0.1317 & 0.2588 & 74.55 & 0.5723 & 0.1703 & 0.2644 \\
PixArt-$\alpha$-XL/2~\cite{chen2023pixart} & $T$ & 85.82 & 0.5220 & 0.1175 & 0.2509 & 158.98 & 0.5261 & 0.1738 & 0.2525 & 103.02 & 0.5996 & 0.1738 & 0.2419 & 74.14 & 0.5778 & 0.1421 & 0.2611 \\
PixArt-$\Sigma$-XL/2~\cite{chen2024pixart} & $T$ & 79.36 & 0.4954 & 0.1574 & 0.2535 & 135.27 & 0.5435 & 0.2271 & 0.2683 & 106.07 & 0.6084 & 0.1139 & 0.2340 & 85.30 & 0.5774 & 0.1476 & 0.2579 \\
\midrule
GLIGEN~\cite{li2023gligen} & $T$ + $B$ & 54.00 & 0.4792 & 0.1658 & 0.2703 & 84.74 & 0.4610 & 0.2109 & 0.2751 & - & - & - & - & - & - & - & - \\
InstanceDiffusion~\cite{wang2024instancediffusion} & $T$ + $B$ & 94.18 & 0.5975 & 0.0957 & 0.2063 & 149.42 & 0.6141 & 0.1203 & 0.2119 & - & - & - & - & - & - & - & - \\
\midrule
VQGAN~\cite{xu2023txt2img} & $T$ & 171.37 & 0.4992 & 0.1482 & 0.2170 & 270.86 & 0.5102 & 0.1747 & 0.1865 & 144.24 & \textbf{0.5180} & 0.1523 & 0.2500 & 141.73 & \textbf{0.4610} & 0.2070 & 0.2487 \\
VQVAE~\cite{xu2023txt2img} & $T$ & 188.46 & 0.5878 & 0.1729 & 0.2451 & 204.13 & 0.5695 & 0.2074 & 0.2450 & 267.30 & 0.6216 & 0.1318 & 0.2563 & 250.83 & 0.5681 & 0.1828 & 0.2534 \\
DiffusionSat~\cite{khanna2023diffusionsat} & $T$ + $L$ & 71.55 & 0.5449 & 0.1615 & 0.2597 & 164.79 & 0.5540 & 0.2166 & 0.2430 & 35.27 & 0.5749 & 0.1299 & 0.2854 & 48.71 & 0.5125 & 0.2023 & 0.2831 \\
GeoSynth~\cite{sastry2024geosynth} & $T$ & 45.59 & 0.5413 & 0.1973 & 0.2478 & 85.80 & 0.4384 & 0.2458 & 0.2521 & 70.55 & 0.5498 & 0.1948 & 0.2832 & 47.84 & 0.4989 & 0.2442 & 0.2699 \\
GeoSynth-OSM~\cite{sastry2024geosynth} & $T$ + $O$ & 53.88 & 0.4298 & 0.1979 & 0.2629 & 110.72 & 0.4374 & 0.2135 & 0.2011 & - & - & - & - & - & - & - & - \\
GeoRSSD~\cite{zhang2024rs5m} & $T$ & 129.23 & 0.5378 & 0.1780 & 0.2692 & 206.74 & 0.5564 & 0.2443 & 0.2549 & 95.28 & 0.5623 & 0.1468 & 0.2842 & 89.95 & 0.5272 & 0.2059 & \textbf{0.2897} \\
CRS-Diff~\cite{tang2024crs} & $T$ & 58.13 & 0.4758 & 0.1855 & 0.2798 & 93.13 & 0.4644 & 0.2306 & 0.2715 & 45.06 & 0.5566 & 0.1074 & 0.2769 & 32.99 & 0.5071 & 0.1882 & 0.2863 \\
Text2Earth$*$~\cite{liu2025text2earth} & $T$ & 25.93 & 0.4269 & 0.2371 & 0.2871 & 49.91 & 0.3926 & 0.3219 & 0.2743 & 75.92 & 0.6109 & 0.1929 & 0.2587 & 51.00 & 0.5372 & 0.2481 & 0.2785 \\
\bottomrule
\cellcolor{gray!25}TerraDiT-XL/2-$\alpha$ & \cellcolor{gray!25}$T$ & 
\cellcolor{gray!25}14.21 & \cellcolor{gray!25}0.3972 & \cellcolor{gray!25}0.2599 & \cellcolor{gray!25}0.2787 & 
\cellcolor{gray!25}20.33 & \cellcolor{gray!25}0.3863 & \cellcolor{gray!25}0.3316 & \cellcolor{gray!25}0.2849 &
\cellcolor{gray!25}33.32 & \cellcolor{gray!25}0.5400 & \cellcolor{gray!25}0.2013 & \cellcolor{gray!25}0.2956 &
\cellcolor{gray!25}\textbf{31.76} & \cellcolor{gray!25}0.4992 & \cellcolor{gray!25}\textbf{0.2680} & \cellcolor{gray!25}0.2764 \\
\cellcolor{gray!25}TerraDiT-XL/2-$\Sigma$ & \cellcolor{gray!25}$T$ + $P$ + $L$ & 
\cellcolor{gray!25}\textbf{12.01} & \cellcolor{gray!25}\textbf{0.3779} & \cellcolor{gray!25}\textbf{0.2740} & \cellcolor{gray!25}\textbf{0.2877} &
\cellcolor{gray!25}\textbf{19.18} & \cellcolor{gray!25}\textbf{0.3652} & \cellcolor{gray!25}\textbf{0.3497} & \cellcolor{gray!25}\textbf{0.2902} &
\cellcolor{gray!25}\textbf{32.11}$^\dagger$ & \cellcolor{gray!25}0.5233$^\dagger$ & \cellcolor{gray!25}\textbf{0.2212}$^\dagger$ & \cellcolor{gray!25}\textbf{0.2967}$^\dagger$ &
\cellcolor{gray!25}- & \cellcolor{gray!25}- & \cellcolor{gray!25}- & \cellcolor{gray!25}- \\
\end{tabular}
}
\caption{\textbf{Quantitative and zero-shot evaluation across four datasets.} Lower FID/LPIPS and higher SSIM/CLIP indicate better performance.  $T$ denotes text input, $O$ denotes OSM raster input, $B$ denotes bounding box input, $L$ denotes geographic location conditioning, and $P$ denotes point prompt conditioning. $*$Text2Earth was trained on the entire Git-10M dataset (including our held-out test sets), which may inflate reported performance for those sections. $^\dagger$Text + Location were used as conditioning. RSICD~\cite{lu2017exploring} only has text conditions available.}
\label{tab:combined_results_all}
\end{table*}

\subsection{Quantitative Evaluations}

In this section, we quantitatively compare the performance of various generative models. We conducted experiments on two held-out in-domain datasets: Git-Rand-15k and Git-Spatial-15k, and two out-of-domain datasets: RSICD~\cite{lu2017exploring} and Functional Map of the World (FMoW)~\cite{christie2018functional}. For each model and dataset, we computed FID, LPIPS, SSIM, and CLIP scores. In Table~\ref{tab:combined_results_all}, we present the performance of our models on Git testing datasets. Models that were not exclusively trained on satellite images exhibited poor FID and LPIPS scores, likely due to a domain shift from their training datasets. However, they demonstrated excellent alignment with textual conditions, as evidenced by high CLIP scores. Some satellite-only generative models, such as Txt2Img-MHN and GeoRSSD, showed poor performance on the datasets, while others, such as Text2Earth, GeoSynth and DiffusionSat, show better performance. Notably, our models demonstrated impressive performance across all metrics and both datasets, surpassing virtually all baselines in both perceptual realism and semantic quality. Importantly, TerraDiT-XL/2-$\Sigma$ outperformed TerraDiT-XL/2-$\alpha$, highlighting the value of point-based controls. 

We also conducted a zero-shot evaluation of the models on the RSICD and FMoW datasets (see Table~\ref{tab:combined_results_all}). These datasets represent a substantial domain shift compared to our training datasets. As anticipated, natural image generative models perform poorly on these datasets. Unsurprisingly, Since CRS-Diff was trained on both datasets, it achieves impressive FID, LPIPS, and CLIP scores across them. DiffusionSat, on the other hand, was exclusively trained on FMoW and demonstrates remarkable performance on that dataset. However, our models outperform all baselines in FID and SSIM, indicating greater structural and distributional realism. 
%
This confirms the ability of our models to effectively represent a diverse range of satellite image distributions.

\subsection{Training-free Inpainting with Points}
To assess TerraDiT-XL/2-$\Sigma$’s ability for high-quality, semantically aligned local generation, we evaluated it on training-free inpainting of urban regions from the Git-Rand-15k test set. Inpainting masks cover 10\% to 60\% of each image, and point prompts are sampled within the masked area from a Poisson distribution whose mean scales with mask size. Competing geospatial models receive global captions of the form “a satellite image of \textit{[point prompts]}.” For fair comparison, geolocation conditioning is excluded during TerraDiT inference, and original pixels are restored outside the edited regions when computing metrics. We report the results of the inpainting experiment in Table~\ref{tab:inpaint}. Our method achieves lower FID and higher SSIM than training-free baselines, indicating superior local semantic alignment and edit quality with minimal annotation effort. Note that Text2Earth was trained on the test dataset, yet our model achieves a superior performance with minimal drop in CLIP scores.

\begin{table}[]
\centering
\resizebox{0.5\linewidth}{!}{
\begin{tabular}{l|ccc}
Method & FID $\downarrow$ & SSIM $\uparrow$ & CLIP $\uparrow$ \\
\toprule
GeoRSSD & 29.79 & 0.6896 & 0.1651 \\
GeoSynth & \underline{21.30} & 0.6908 & 0.1684 \\
Text2Earth &  66.20 & \underline{0.6987} & \textbf{0.1768} \\
\bottomrule
\cellcolor{gray!25}TerraDiT-XL/2-$\Sigma$ & \cellcolor{gray!25}\textbf{16.81} & \cellcolor{gray!25}\textbf{0.7000} & \cellcolor{gray!25}\underline{0.1738}
\end{tabular}}
\caption{\textbf{Training-free Inpainting with Points}. TerraDiT performs well on the task of inpainting as compared to other geospatial generative models.}
\label{tab:inpaint}
\end{table}

\subsection{Ablations}
To evaluate the components of TerraDiT, we conduct several ablations. For each setting, we train the models from scratch up to 400k training steps. We then evaluate each of the trained models on the Git-Rand-15k test set. Since our goal in these ablations is to assess the model’s ability to learn the overall distribution of the Git-10M dataset, we avoid using the spatial hold out because it is limited in geolocation and would therefore obscure how much the model benefits from geolocation information. Please see the appendix for additional ablations on the choice of geolocation and satellite imagery encoders during training.

\begin{table}[]
\centering
\resizebox{0.5\linewidth}{!}{
\begin{tabular}{l|c|ccc}
Iter.& Conditions & FID $\downarrow$ & SSIM $\uparrow$ & CLIP Score $\uparrow$ \\
\toprule
400k&-&37.97&0.2111&0.2423\\
\midrule
400k&Text&35.57&0.2397&0.2587\\
400k&Geolocation&37.83&0.2132&0.2522\\
400k&Points&35.65&0.2529&0.2604\\
\midrule
400k&Text + Geolocation&31.72&0.2513&0.2636\\
400k&Text + Points&31.26&0.2567&0.2678\\
\bottomrule
\cellcolor{gray!25}400k&\cellcolor{gray!25}Text + Points + Geolocation&\cellcolor{gray!25}\textbf{28.34}&\cellcolor{gray!25}\textbf{0.2589}&\cellcolor{gray!25}\textbf{0.2699}\\
\end{tabular}}
\caption{Choice of input modalities and corresponding generative performance of TerraDiT.}
\label{tab:inp_mod}
\end{table}

\begin{table}[]
\centering
\resizebox{0.5\linewidth}{!}{
\begin{tabular}{l|c|ccc}
Iter.& Conditions & FID $\downarrow$ & SSIM $\uparrow$ & CLIP Score $\uparrow$ \\
\toprule
400k&Cross Attention&31.15&0.2429&0.2591\\
\bottomrule
\cellcolor{gray!25}400k&\cellcolor{gray!25}ALA&\cellcolor{gray!25}\textbf{28.34}&\cellcolor{gray!25}\textbf{0.2589}&\cellcolor{gray!25}\textbf{0.2699}\\
\end{tabular}}
\caption{Choice of attention module for point queries.}
\label{tab:asla_choice}
\end{table}

Firstly, we evaluate the impact of various combinations of conditioning modalities in Table~\ref{tab:inp_mod}. As shown, adding geolocation and point queries both enhance the generation performance of TerraDiT models. Adding geolocation improves FID scores, though it has minimal impact on SSIM and CLIP scores. In contrast, adding point queries substantially boosts the semantic and perceptual quality of generated images, highlighting their critical role in controlling and improving image synthesis. We also conduct an experiment to evaluate the effectiveness of our proposed ALA block. In Table~\ref{tab:asla_choice} we compare it by removing the MetaRBF module and only retaining the vanilla cross-attention module. We see that our ALA block outperforms vanilla cross-attention mechanism. It effectively captures the spatial extent of various concepts based on their semantic meaning and structure in the real world (see Figure~\ref{fig:asla_viz}).

\section{Conclusions}
In this work, we introduced TerraDiT, a point-conditioned diffusion transformer designed to be annotation-friendly for satellite image generation. We proposed a novel adaptive local attention module that effectively conditions the generation process using point queries. This module adaptively regularizes attention scores based on the spatial location and semantic information of the point queries. Our model was evaluated on four in-domain and out-of-domain datasets, and the results show that it outperforms state-of-the-art generative models in both perceptual and semantic quality. Additionally, our point-conditioned model demonstrates its ability to effectively capture the structure imposed by the point queries. Future work will focus on incorporating other types of input conditions, such as lines or polygons, to make our models more flexible and scalable to diverse applications. 


%
%
\section*{Acknowledgements}
This research used the TGI RAILs advanced compute and data resource which is supported by the National Science Foundation (award OAC-2232860) and the Taylor Geospatial Institute.
\clearpage
\bibliographystyle{splncs04}
\bibliography{main}
\clearpage
\section*{\centering Supplementary Material}
\appendix
\renewcommand{\theHsection}{appendix.\thesection}
\renewcommand{\theHsubsection}{appendix.\thesection.\arabic{subsection}}

\section{Encoder Ablation Studies}
In this section, we present ablations on the satellite image encoders used for representation alignment during training and the geolocation encoders used for geolocation conditioning.
\subsection{Representation Alignment}
We evaluate our representation alignment framework by comparing several satellite image encoders as target representations, following the REPA framework~\cite{yu2024representation}. We learn an MLP that maps diffusion latents into the representation space of a fixed, pretrained encoder, enabling the DiT backbone to inherit informative spatial features early in training.

\begin{table}[!b]
\centering
\resizebox{0.5\linewidth}{!}{
\begin{tabular}{l|c|ccc}
Iter.& Target Repr. & FID $\downarrow$ & SSIM $\uparrow$ & CLIP Score $\uparrow$ \\
\toprule
400k&None &48.07 & 0.2428 & 0.2557\\
\midrule
400k&RemoteCLIP~\cite{liu2024remoteclip} &48.22&0.2689&\textbf{0.2636}\\
400k&Satlas~\cite{bastani2023satlaspretrain} &40.14&0.2703&0.2588\\
400k&SatMAE-L~\cite{cong2022satmae} &41.61&0.2750&0.2633\\
400k&ScaleMAE-L~\cite{reed2023scale} &50.89&0.2689&0.2635\\
400k&SatMAE++-L~\cite{noman2024rethinking} &45.20&0.2747&0.2618\\
400k&AnySat~\cite{astruc2025anysat} &43.21&0.2694&0.2543\\
\midrule
\midrule
400k&DINOv3-L~\cite{simeoni2025dinov3} &\textbf{35.57}&\textbf{0.2763}&0.2617\\
\midrule
\end{tabular}}
\caption{Choice of satellite image representations to align with diffusion transformer. We follow REPA~\cite{yu2024representation} and align the hidden representations of 8th TerraDiT block with the satellite image representations.}
\label{tab:satrepa}
\end{table}

As shown in Table~\ref{tab:satrepa}, we train each variant for 400k steps on the Git-10M dataset using our TerraDiT-B/2 model, and report FID, SSIM, and CLIPScore. While representation alignment aims to improve training efficiency, we find that the choice of target encoder is crucial, as some satellite image encoders degrade performance relative to the no-REPA baseline. Among different satellite image encoders, the satellite image only DINOv3-L image encoder achieves the best FID and SSIM while maintaining competitive CLIPScore. These results indicate that DINOv3-L provides the most effective alignment signal during training, and thus we adopt it in other experiments. 

\subsection{Geolocation Conditioning}

\begin{table}[t]
\centering
\resizebox{0.5\linewidth}{!}{
\begin{tabular}{l|c|ccc}
Iter.& Geo Emb. & FID $\downarrow$ & SSIM $\uparrow$ & CLIP Score $\uparrow$ \\
\toprule
400k&None &35.57&0.2397&0.2587 \\
\midrule
400k&SINR~\cite{cole2023spatial} &34.56&0.2366&0.2587\\
400k&GeoCLIP~\cite{vivanco2023geoclip} &33.44&0.2389&0.2615\\
400k&TaxaBind~\cite{sastry2025taxabind} &37.24&0.2480&0.2591\\
400k&SatCLIP~\cite{klemmer2025satclip} &36.03&0.2472&0.2606\\
400k&Climplicit~\cite{dollinger2025climplicit} &37.13&0.2424&0.2592\\
\midrule
\midrule
400k&RANGE~\cite{dhakal2025range} &\textbf{31.72}&\textbf{0.2513}&\textbf{0.2636}\\
\midrule
\end{tabular}}
\caption{Choice of location representations for conditioning.}
\label{tab:geoloc}
\end{table}

As seen in prior works~\cite{khanna2023diffusionsat, sastry2024geosynth}, incorporating geolocation cues can significantly improve the realism and faithfulness of generated satellite imagery.
We similarly incorporate geolocation into TerraDiT and compare several embedding strategies. As shown in Table~\ref{tab:geoloc} some encoder choices outperform the no geolocation conditioning baseline, while others can degrade performance. We observe the strongest improvements using RANGE~\cite{dhakal2025range} as our geolocation encoder, achieving the best FID, SSIM, and CLIPScore. This suggests that RANGE offers a more expressive and diffusion-friendly geolocation prior for satellite image generation.

\section{Implementation Details}
\subsection{Training}
We follow the training setup used by REPA~\cite{yu2024representation}. We primarily use the Git-10M dataset for training. This dataset contains satellite images at varying zoom levels. For simplicity, we focus on zoom level 17, which has a ground sampling distance of approximately 1 meter per pixel. We leave exploring multiple zoom levels as future work. Each image in the resulting dataset has a resolution of 256x256. We employ the latent space of SDXL VAE~\cite{podell2023sdxl} for training our diffusion transformer. The encoder of the VAE performs an 8x downsampling of input images, resulting in each image in our dataset being converted into a latent of size 32x32. To accelerate training, we precompute the latents for each image in our training dataset. All our experiments utilize a patch size of 2. Consequently, the hidden latent input for our model has size 16x16. We employ LongCLIP~\cite{zhang2024long} for encoding the image-level and point-level textual descriptions. For comprehensive details about the architecture and training settings, refer to Table~\ref{tab:hyperparams}.

\subsection{Point Prompt Generation}
As discussed earlier, we expanded the Git-10M dataset by downloading OSM vector annotations for each image. Each OSM vector annotation includes global geolocation information, which we convert to image pixel coordinates (0-255) since each image is 256x256. During training, we randomly select pixel coordinates that have some OSM annotation. For each point, we extract OSM tags from the corresponding vector annotations. In certain cases, a pixel coordinate may be associated with multiple OSM tags. In such instances, we randomly select one tag among the possible tags. Note that the input to our model is a latent of size 16x16. Consequently, we transform the pixel coordinates for each point prompt by dividing them by 16. 

\begin{table}
\centering
\begin{tabular}{l|l}
Hyperparameter & Value\\
\midrule
input\_dim&32x32x4\\
patch\_size&2\\
num\_layers&28\\
hidden\_dim&1152\\
num\_head&16\\
\midrule
num\_gpus&4\\
gpu\_type&NVIDIA H100\\
optimizer&AdamW\\
lr&1e-5\\
batch\_size&256\\
num\_points$^{\dag}$& randint(0, 50)\\
\end{tabular}
\caption{Architectural and Hyperparameter settings for training TerraDiT. $^{\dag}$Only for TerraDiT-XL/2-$\Sigma$.}
\label{tab:hyperparams}
\end{table}

\subsection{Inference}
 \textbf{Generation settings.} For all our TerraDiT variants, we follow the same inference procedure to generate images for evaluation. For all our TerraDiT variants, we follow the same inference procedure to generate images for evaluation. All details are provided in Table~\ref{tab:gen_params}. We use 100 flow steps with a batch size of 16. For our TerraDiT-XL/2-$\Sigma$, we sample 20-50 points which are selected uniformly at random. 

\textbf{Training-Free Inpainting.} We perform training-free inpainting using a procedure analogous to the Stable Diffusion inpainting pipeline~\cite{podell2023sdxl}. At inference time, we keep the unmasked region fixed in latent space by encoding it with the SDXL-VAE and adding noise according to the forward diffusion schedule at each timestep. The denoising model is then applied to the full latent, but only the masked region is updated across steps, effectively restricting generation to the missing content while preserving the context. All sampling hyperparameters follow those specified in Table~\ref{tab:hyperparams}.

\begin{table}
\centering
\begin{tabular}{l|l}
Hyperparameter & Value\\
\midrule
num\_gpus&1\\
num\_steps&100\\
sampler\_type&ode\\
cfg\_scale&0.0\\
batch\_size&16\\
num\_points$^{\dag}$& randint(20, 50)\\
\end{tabular}
\caption{Hyperparameter settings for generating satellite images for each experiment. $^{\dag}$Only for TerraDiT-XL/2-$\Sigma$.}
\label{tab:gen_params}
\end{table}

\begin{figure}
    \centering
    \includegraphics[width=0.75\linewidth]{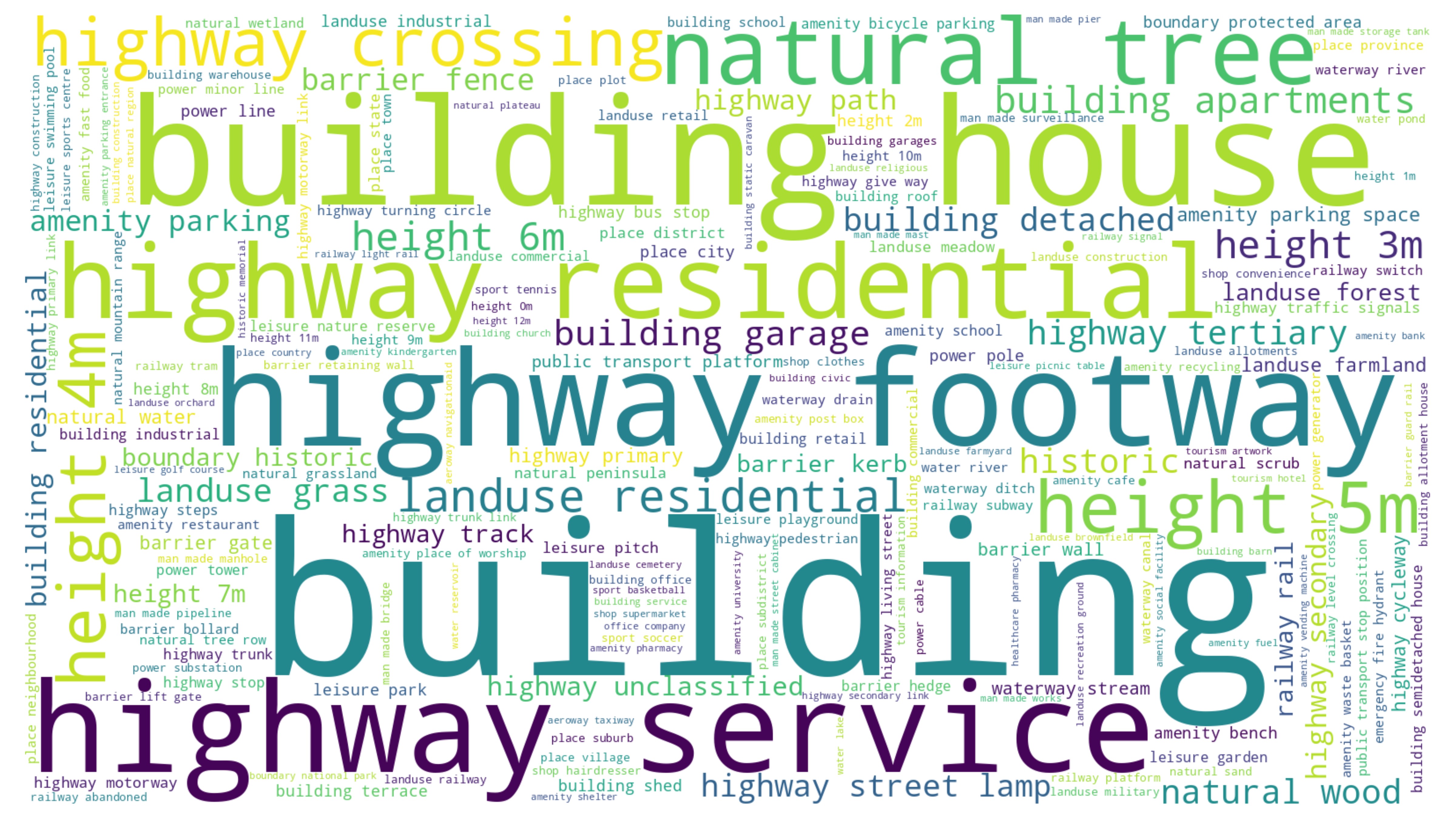}
    \caption{Wordcloud of semantic tags used for point queries. Larger words indicate more common tags.}
    \label{fig:tag_wordcloud}
\end{figure}

\begin{figure*}[!t]
  \centering
  \begin{subfigure}[b]{0.32\textwidth}
    \centering
    \includegraphics[width=\linewidth]{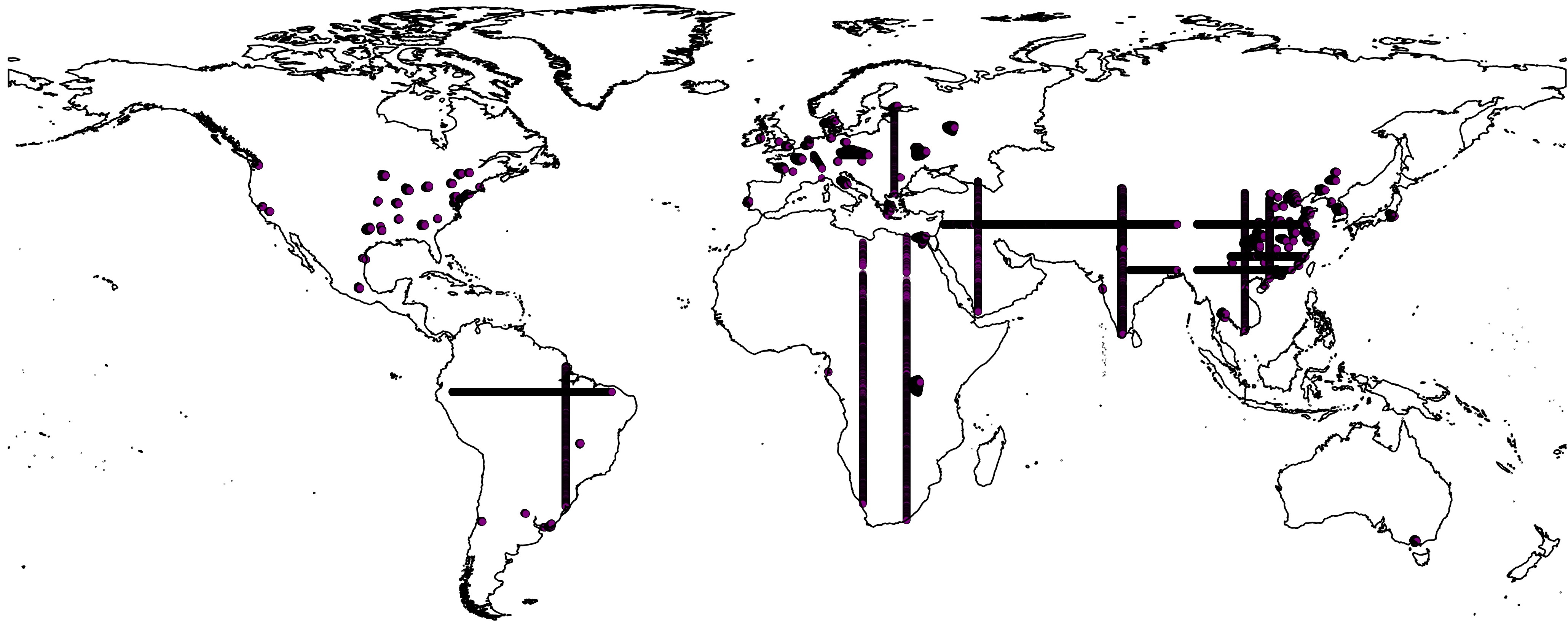}
    \caption{Training Data}
    \label{fig:train}
  \end{subfigure}\hfill
  \begin{subfigure}[b]{0.32\textwidth}
    \centering
    \includegraphics[width=\linewidth]{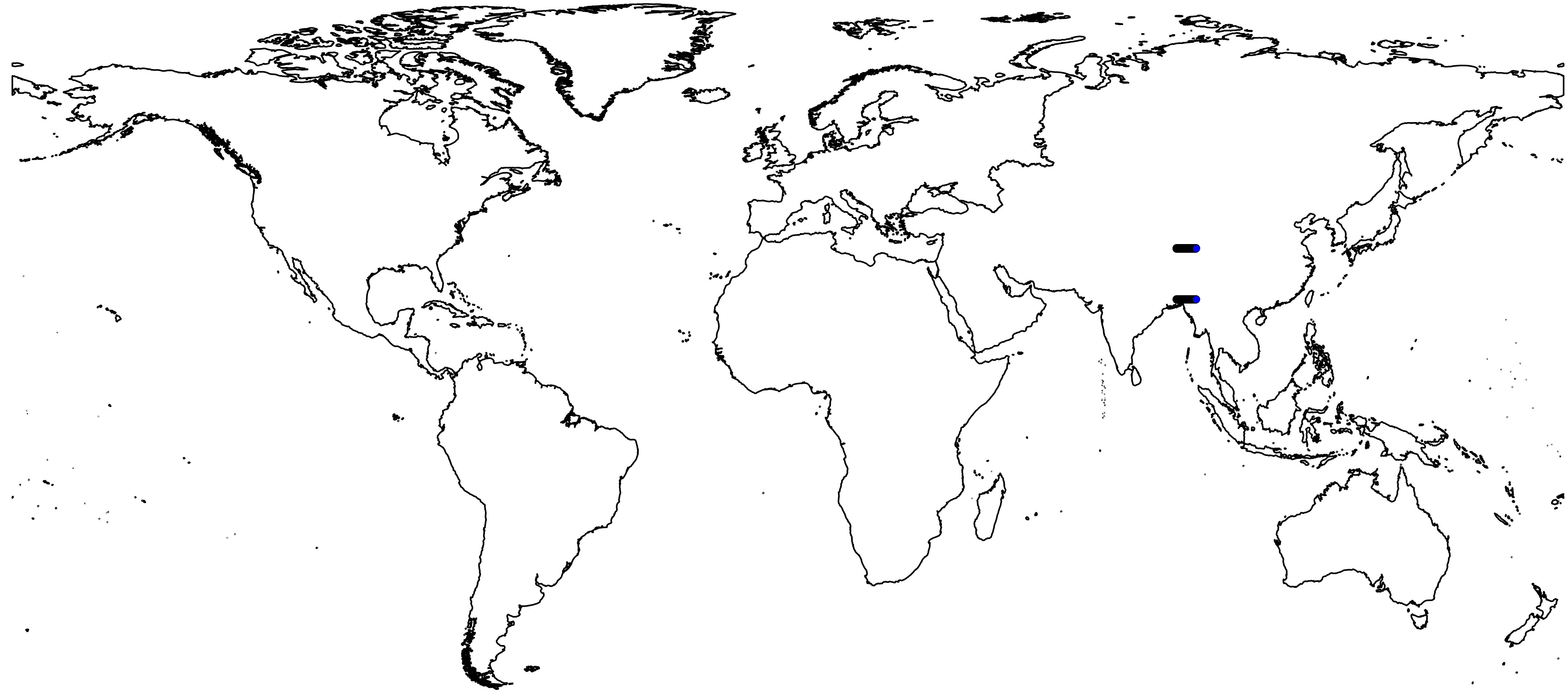}
    \caption{Git-Spatial-15k}
    \label{fig:spatial}
  \end{subfigure}\hfill
  \begin{subfigure}[b]{0.32\textwidth}
    \centering
    \includegraphics[width=\linewidth]{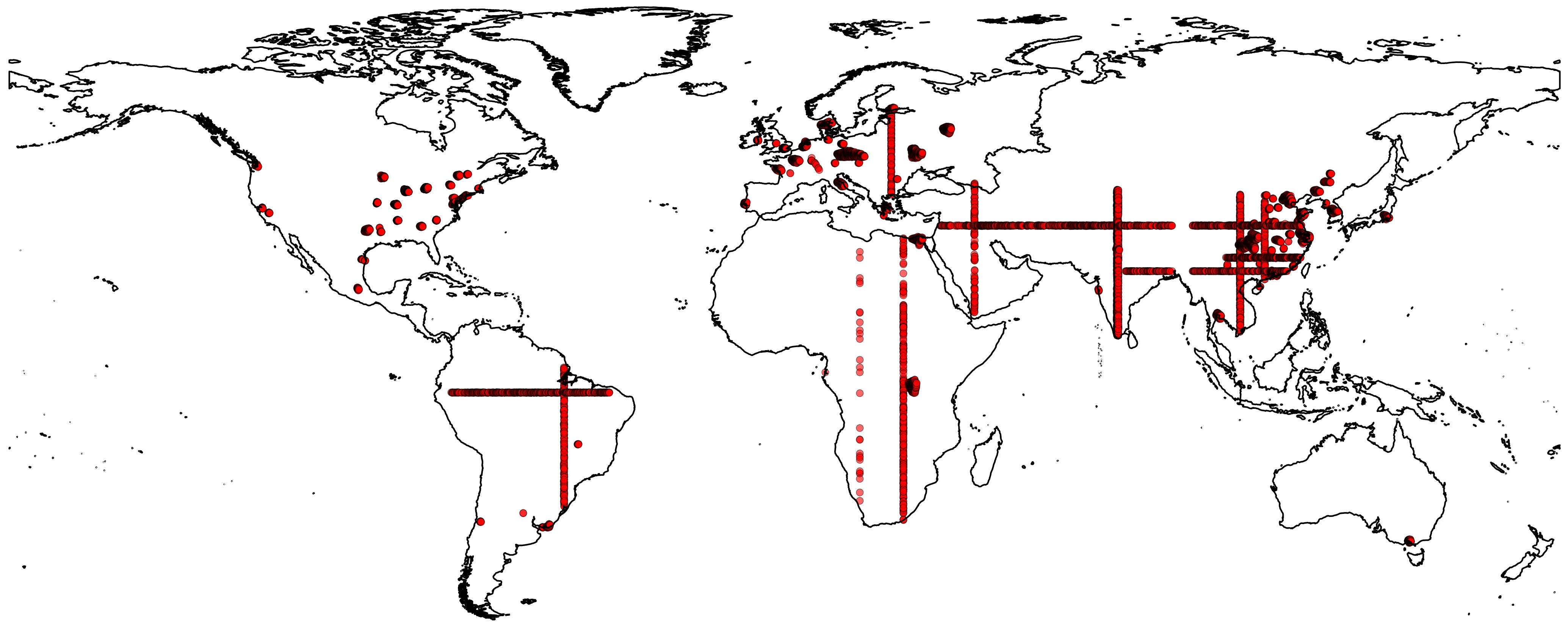}
    \caption{Git-Random-15k}
    \label{fig:random}
  \end{subfigure}

  \vspace{0.5em}
  \caption{Training and test splits. (a) training data, (b) spatial holdout test set, (c) random holdout test set.}
  \label{fig:three-maps}
\end{figure*}

\begin{table*}[t]
\centering
\begin{tabular}{l|c|cccc}
Model      & Condition & \textit{Git-Rand-15k} $\downarrow$ & \textit{Git-Spatial-15k} $\downarrow$ & RSICD $\downarrow$ & FMoW $\downarrow$\\
\toprule
SDXL~\cite{podell2023sdxl} & $T$ &92.94&94.48&98.86&97.05\\
SD 3~\cite{esser2024scaling} &$T$&96.22&97.14&97.40&98.26\\
PixArt-$\alpha$-XL/2~\cite{chen2023pixart} & $T$ &98.83&98.66&99.12&99.39\\
PixArt-$\Sigma$-XL/2~\cite{chen2024pixart} & $T$ &97.03&96.73&99.00&99.37 \\
\midrule
\midrule
DiffusionSat~\cite{khanna2023diffusionsat} & $T + L$ & 95.04& 99.19&93.36&94.68\\
GeoSynth~\cite{sastry2024geosynth}   & $T$      &99.81&99.83&99.44&99.80\\
GeoSynth-OSM~\cite{sastry2024geosynth}   & $T + O$  & 99.68&99.84&-&-\\
VQGAN~\cite{xu2023txt2img}& $T$  &97.08 &98.01 &\underline{91.94} &98.56\\
VQVAE~\cite{xu2023txt2img}& $T$  &99.95&100.00&99.95&99.88\\
GeoRSSD~\cite{zhang2024rs5m}& $T$   &95.24 &97.60&96.61&97.18\\
CRS-Diff~\cite{tang2024crs}   & $T$  & 96.55 &98.25 & 99.53 & 96.98  \\
Text2Earth~\cite{liu2025text2earth} & $T$   & 77.93&\underline{80.31}&93.86&\textbf{92.15}\\
\midrule
\midrule
TerraDiT-XL/2-$\alpha$     & $T$         &  \underline{74.25}&\textbf{78.22}&\textbf{91.85}&\underline{92.35}\\
TerraDiT-XL/2-$\Sigma$   & $T + P + L$    &     \textbf{72.66}  & 81.70 &-&-\\
\end{tabular}
\caption{AI-generated image detection using the held-out test sets and generated images. We report F1 scores achieved by the detector for the fake class. Lower scores indicate better performance of the models.}
\label{tab:deepfake}
\end{table*}

\section{Dataset}
For completeness, we include visualizations of the full dataset distribution (see Figure~\ref{fig:three-maps}). These figures show the global coverage of the training set and the two evaluation splits. The spatial and random holdouts are presented alongside the training data to provide a visual sense of geographic diversity and sampling strategy. In addition, we visualize the distribution of semantic tags used in point queries by rendering a wordcloud of tag frequencies (Figure~\ref{fig:tag_wordcloud}). 

\section{AI-generated Image Detection}
To evaluate the realism and perceptual quality of images generated by different generative models, we train an AI-generated image detector for each model. For each model and held-out test set, we combine the generated and ground truth images. Subsequently, we randomly split this combined data into training, validation, and testing sets with a ratio of 0.8:0.05:0.15. For each model and test set, we independently train a simple neural network consisting of a single convolution layer, a max pooling layer, and a linear classification layer. We use a batch size of 128, Adam optimizer, and a learning rate of 3e-4. Each model is trained for a maximum of 15 epochs, and the model with the best performance on the validation data is used for final comparison. In Table~\ref{tab:deepfake}, we report the F1 scores for the fake image class achieved for each training run. Lower F1 scores indicate better performance of the generative models.

\section{More Qualitative Examples}
\begin{figure*}[p]
  \centering
  \includegraphics[width=\textwidth]{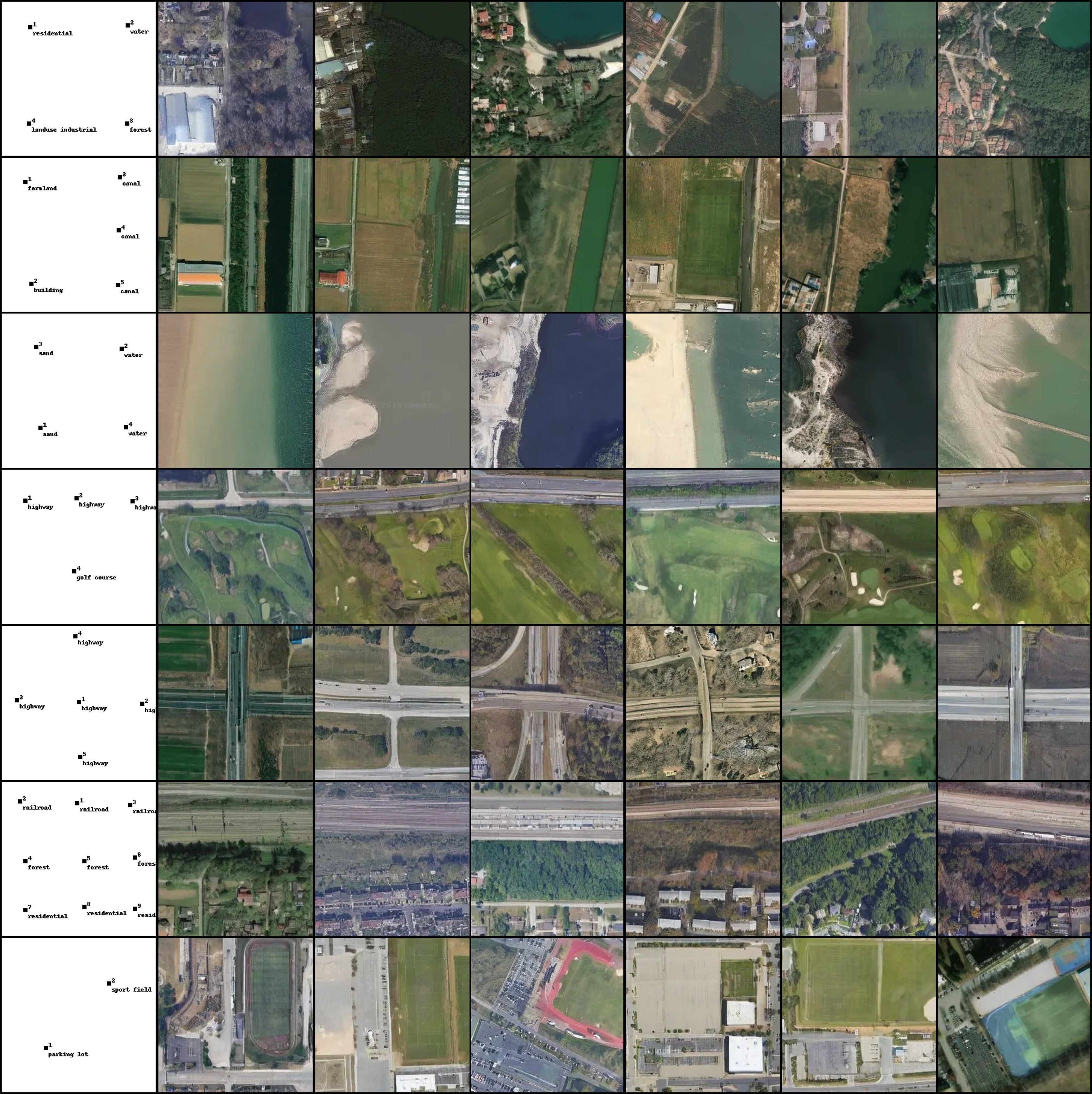}
  \caption{Additional qualitative examples generated using TerraDiT-XL/2-$\Sigma$.}
  \label{fig:geodit_point_prompts}
\end{figure*}

\begin{figure*}[p]
  \centering
  \includegraphics[width=\linewidth]{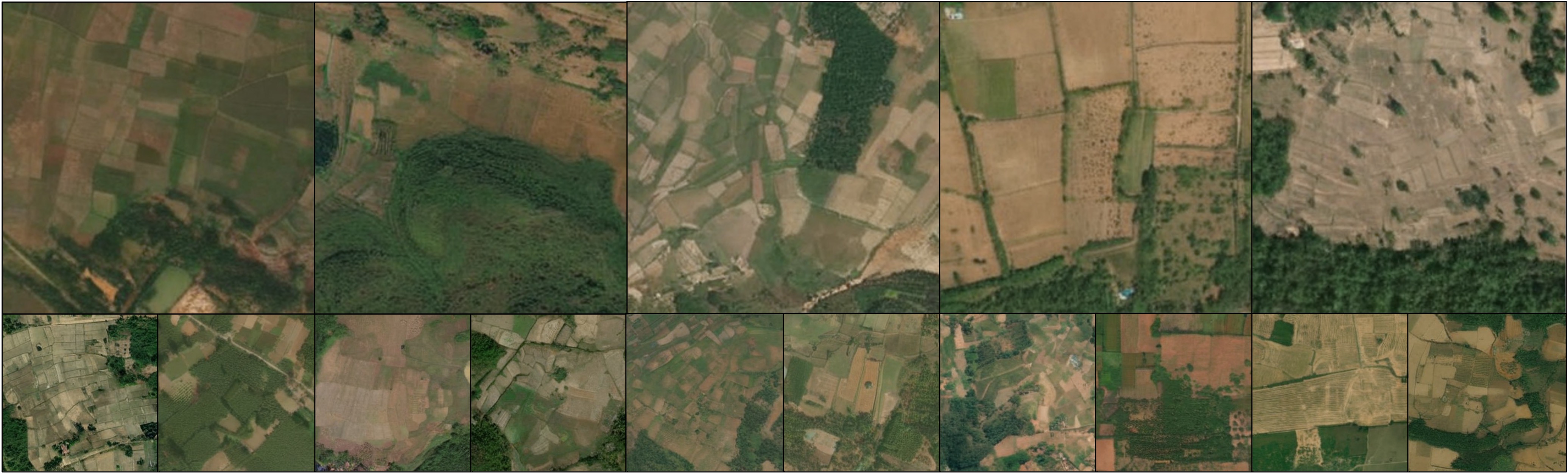}
  \caption{Caption: ``This satellite image shows a rural landscape with a patchwork of agricultural fields. The fields vary in size and shape, some of which appear to be recently harvested or fallow. There is a cluster of dense green vegetation at the bottom of the image, suggesting a forested area."}
  \label{fig:collage1}
\end{figure*}

\begin{figure*}[p]
  \centering
  \includegraphics[width=\linewidth]{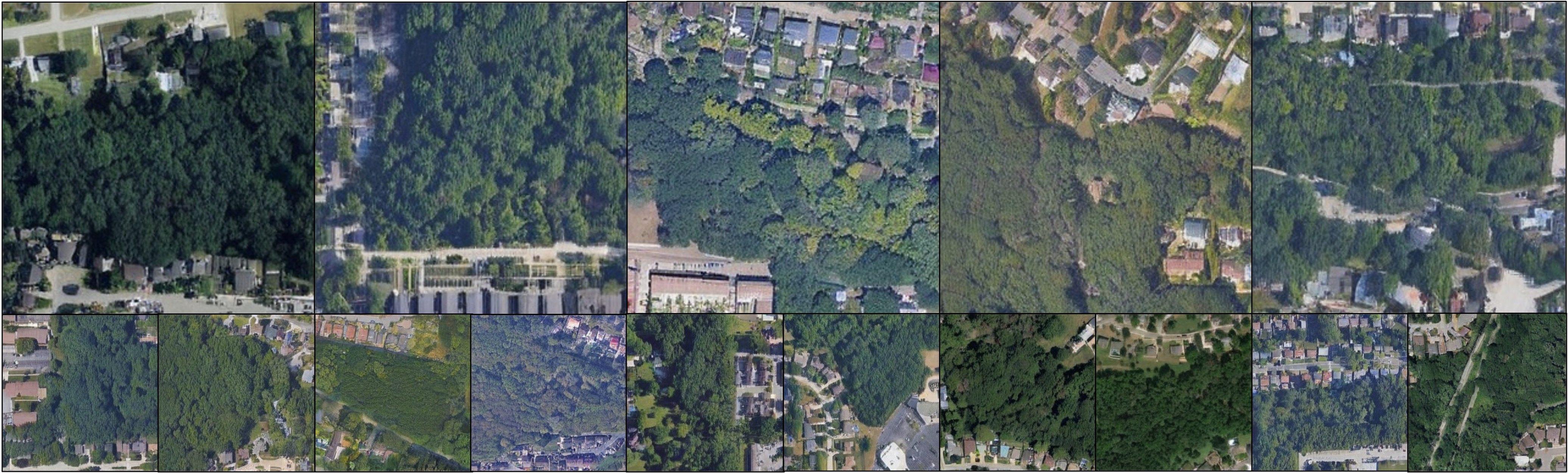}
  \caption{Caption: ``This satellite image depicts a residential area adjacent to a forested region. The top part of the image shows a row of houses with visible rooftops and parked vehicles. The middle section consists of a dense forest with a variety of trees. The bottom part includes additional houses and a swimming pool within a residential neighborhood."}
  \label{fig:collage2}
\end{figure*}

\begin{figure*}[p]
  \centering
  \includegraphics[width=\linewidth]{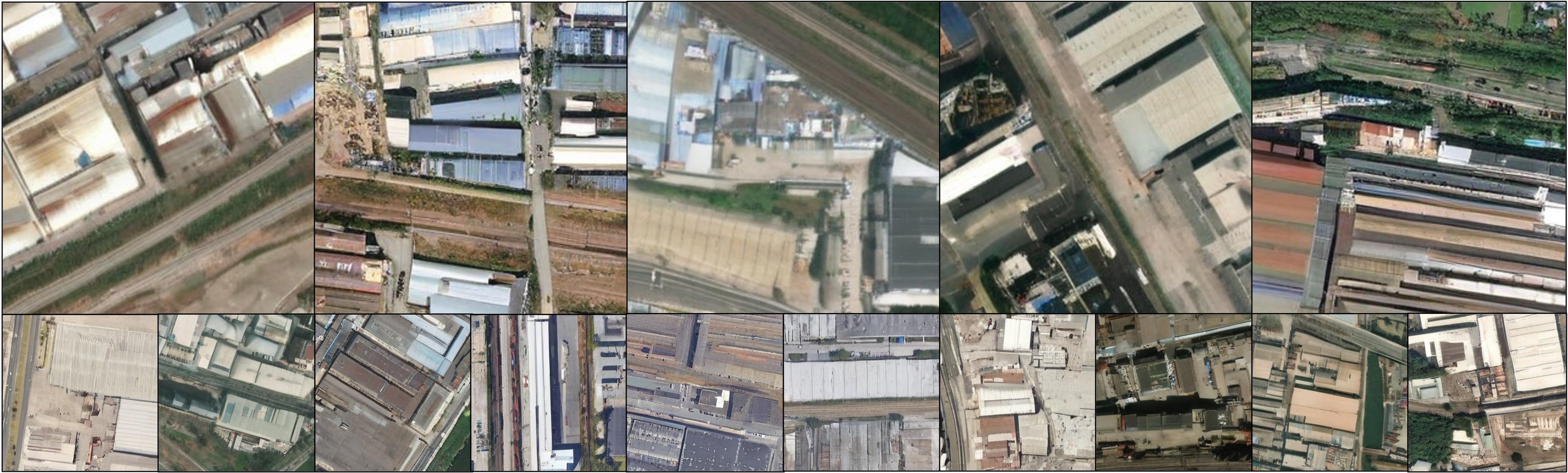}
  \caption{Caption: ``The satellite image shows a large industrial area with multiple warehouse buildings. One building has a section of its roof that is partially covered in a different material or color. Adjacent to the buildings, there is a road running parallel to a set of railway tracks. The area appears to be an industrial or commercial zone with infrastructure for transportation and storage."}
  \label{fig:collage3}
\end{figure*}

\begin{figure*}[p]
  \centering
  \includegraphics[width=\linewidth]{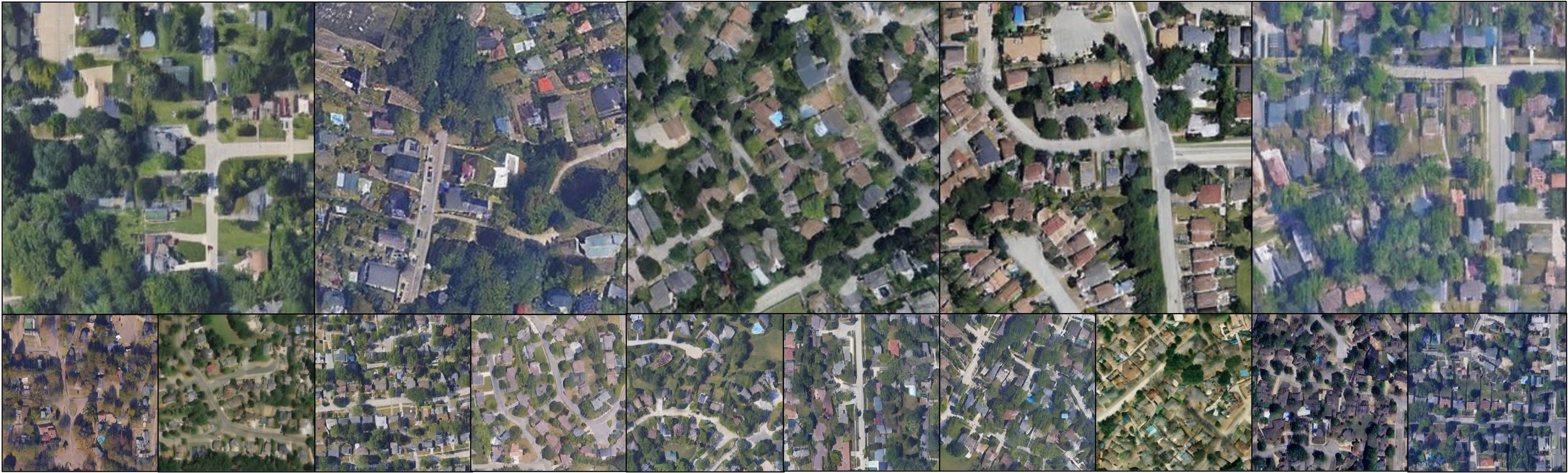}
  \caption{Caption: ``The satellite image shows a residential neighborhood with houses, trees, and roads. The layout includes curved streets and intersections. Several houses have visible driveways and backyards, some of which contain blue-colored objects, likely swimming pools. The area is densely populated with trees and greenery."}
  \label{fig:collage4}
\end{figure*}

\begin{figure*}[p]
  \centering
  \includegraphics[width=\linewidth]{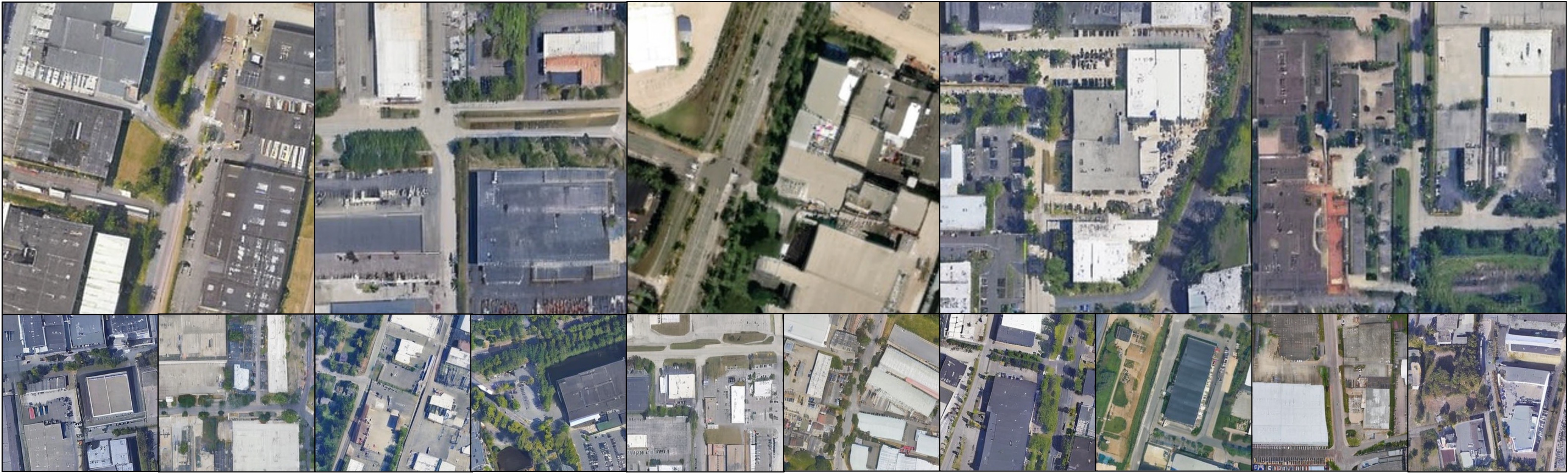}
  \caption{Caption: ``The satellite image shows an industrial area with several large buildings and adjacent parking lots. There are multiple vehicles parked around the buildings and along the streets. Trees and greenery are present in the vicinity, particularly towards the top right corner of the image. The layout includes straight roads separating the buildings and parking areas."}
  \label{fig:collage5}
\end{figure*}

\begin{figure*}[p]
  \centering
  \includegraphics[width=\linewidth]{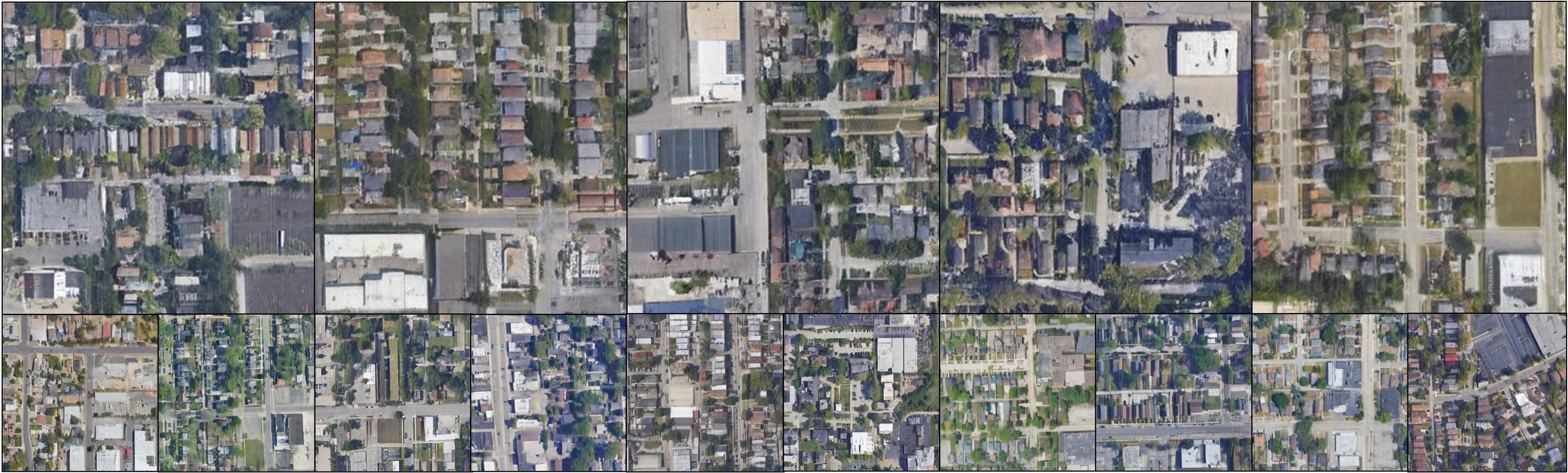}
  \caption{Caption: ``The satellite image shows an urban residential area with grid-like streets. There are numerous houses, mostly single-family homes, arranged in neat rows along the streets. At the center, there is a large industrial or commercial building with a sizable parking lot adjacent to it. The surrounding area includes some trees and patches of greenery."}
  \label{fig:collage6}
\end{figure*}

We illustrate the diverse visuals of our TerraDiT-XL/2-$\alpha$ with generated collages in Figures~\ref{fig:collage1}--\ref{fig:collage6}. The $\alpha$ model captures a wide variety of geographic contexts, including rural, suburban, urban, and industrial zones. Complementing this imagery, we show TerraDiT-XL/2-$\Sigma$ generated imagery from diverse point prompts in Figure~\ref{fig:geodit_point_prompts}, highlighting its ability to produce high-fidelity samples conditioned on various point queries.

\vspace{-5mm}
\end{document}